\documentclass[fleqn,10pt]{wlscirep}
\usepackage[utf8]{inputenc}
\usepackage[T1]{fontenc}
\usepackage{bm}
\usepackage{hyperref}
\usepackage{multirow}
\usepackage{graphicx}
\usepackage{siunitx}
\usepackage{acro}

\title{Eversion-based robots can enable safe access, steering and endoscopic imaging within the spinal subarachnoid space}

\author[1,*]{Zicong Wu}
\author[2]{Panagiotis Kalozoumis}
\author[3]{S.M.Hadi Sadati}
\author[4]{Aminul I. Ahmed}
\author[1]{Jonathan Shapey}
\author[1]{Christian Baker}
\author[1]{Thomas Booth}
\author[1]{Wenfeng Xia}
\author[1]{Sebastien Ourselin}
\author[5]{Panagiotis Vartholomeos}
\author[1]{Christos Bergeles}
\affil[1]{Department of Surgical \& Interventional Engineering, School of Biomedical Engineering \& Imaging Sciences, Faculty of Life Sciences \& Medicine, King's College London, London WC2R 2LS, United Kingdom}
\affil[2]{Department of Computer Science \& Biomedical Informatics, University of Thessaly, Lamia 35131, Greece}
\affil[3]{School of Engineering and Materials Science, Queen Mary University London, London E1 4NS, United Kingdom}
\affil[4]{Institute of Psychiatry, Psychology \& Neuroscience, King's College London, London WC2R 2LS, United Kingdom}
\affil[5]{School of Electrical and Computer Engineering, National Technical University of Athens, Zografou 15773, Greece}
\affil[*]{Corresponding Author: zicong.wu@kcl.ac.uk}

\begin{abstract}
Safe navigation within the spinal subarachnoid space is constrained by its narrow, compliant, and delicate anatomy. Conventional catheters and continuum robots rely on proximal pushing, thereby propagating frictional and shear forces along the tissue-device interface, limiting distal controllability and increasing the risk of neural injury. Here, we demonstrate a robotic platform enabling ``friction-less'' controlled extension and steering within the human spinal subarachnoid space, validated through computational modelling, phantom experiments and intact human cadaver studies. The proposed 2 mm-diameter eversion-growing robot integrates a miniature endoscope for real-time intrathecal visualisation and advances through pressure-driven tip eversion, localising motion to the distal tip while minimising translational sliding of its deployed body. 
Phantom experiments demonstrated reductions of $65.2\%$ in mean interaction force and $48.0\%$ in peak interaction force compared with matched push-based insertion. 
Complementary physics-based modelling further revealed that eversion-based growth redistributes tissue loading, reducing local stress concentrations and interfacial shear compared with conventional push-based insertion.
A controlled intrathecal extension of $150\,$mm with concurrent fluoroscopic and in situ endoscopic visualisation was achieved in an intact human cadaver, providing clinically relevant access across multiple vertebral levels from a standard lumbar entry point. 
Post-procedural anatomical inspection following laminectomy and durotomy revealed no observable macroscopic disruption of the dura mater or surrounding neural structures. Together, these findings establish the first mechanically characterised and multimodally validated demonstration of eversion-based robotic navigation in intact human spinal anatomy, providing a quantitative and procedural foundation for future intrathecal interventions. Further validation across larger anatomical cohorts and under physiological conditions will be required before clinical translation.
\end{abstract}

\begin{document}

\flushbottom
\maketitle

\thispagestyle{empty}

\section*{Introduction}
Neurological disorders are the leading cause of disability and the second leading cause of death worldwide, affecting over $40\%$ of the global population, with increasing predicted prevalence as populations age and life expectancy increases \cite{wang2024expanding}. Many of these conditions, particularly chronic pain and severe spasticity, cannot be adequately relieved with systemic therapies alone. Intrathecal drug delivery introduces therapeutics directly into the cerebrospinal fluid (CSF) within the subarachnoid space, bypassing the blood-brain barrier and enabling effective target engagement with lower medication doses \cite{de2022intrathecal,england2016clinical}. Despite recommendations and clinical evidence of effectiveness, intrathecal provision of drugs remains limited relative to the estimated eligible patient population \cite{eldabe2024intrathecal}. For example, in England, approximately 450 patients were considered for intrathecal drug delivery for cancer pain, despite an estimation of at least $8,000$ eligible cases (2020 data\cite{duarte2020unmet}). This disparity underscores the unmet clinical need for safer and more controllable intrathecal interventions.

Safe and effective intrathecal access, however, remains technically challenging because the spinal subarachnoid space is a narrow, compliant, fluid-filled compartment bounded by the dura mater, arachnoid mater, and pia mater with opposed nerve roots, connective trabeculae, and other fine anatomical structures. 
At the lumbar level, the dural sac typically measures approximately $12$–$18\,$mm in transverse diameter and $10$–$15\,$mm in the anteroposterior direction \cite{lee2021extent,pierro2017sagittal}. However, these gross dimensions substantially overestimate the effective intrathecal navigation corridor. Anatomical measurements indicate that the lateral subarachnoid space between the spinal cord and the dura averages approximately $2.5\,$mm per side \cite{zaaroor2006morphological}, while the dural sac thickness itself is approximately  $0.307\pm0.122\,$mm \cite{hong2011analysis}. Therefore, although standard clinical access methods, such as lumbar puncture, are widely used to sample CSF or administer drugs, the difficulty of intrathecal navigation can vary substantially with patient anatomy, spinal level, and pathological changes. This makes deeper navigation beyond the initial lumbar entry technically demanding \cite{kadamus2025multimodal}.


\begin{figure*}[!t]
    \centering
    \includegraphics[width = 0.85\textwidth]{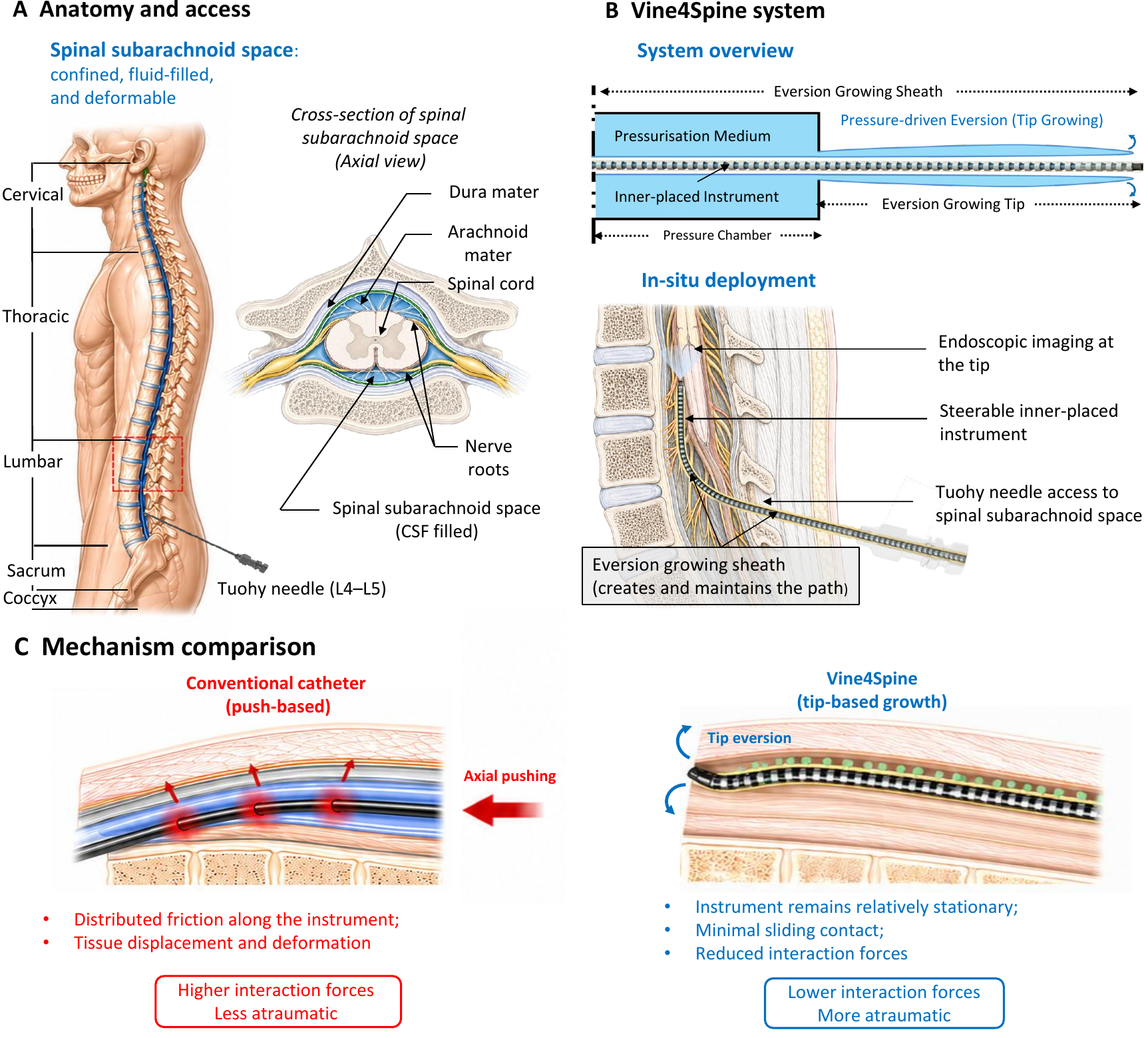}
    \caption{Vine4Spine and growing-based navigation in spinal subarachnoid space. \textbf{(A)} The anatomy of the spinal subarachnoid space is confined, fragile, deformable, and CSF-filled. \textbf{(B)} System overview of Vine4Spine and its in-situ deployment in the spinal subarachnoid space via pressure-driven tip-eversion-based growing. \textbf{(C)} Compared with conventional push-based catheters, Vine4Spine is deployed via tip-eversion-based growing. Friction and shear forces are suppressed due to non-translational motion from eversion-based navigation.}
    \label{fig.conception}
\end{figure*}

The challenges behind elongated and controllable navigation stem from the fundamental mechanics of conventional push-based catheters and guidewires, which remain the clinical mainstay for intraluminal procedures. These devices transmit axial loads from their proximal base to their distal tip, propagating frictional interaction forces along their shaft as they advance. In confined and compliant anatomies such as the spinal subarachnoid space, accumulated friction and axial loading reduce distal tip controllability and increase the risk of iatrogenic injury \cite{mannan2023current,pozza2023spinal}.  

Robotics researchers have sought to address the access and navigation challenges in confined and delicate anatomical environments through compliant continuum robot designs \cite{burgner2015continuum,majidi2014soft,suulker2026state}. While such systems can improve flexibility and steerability, they fundamentally rely on base-driven insertion and therefore retain the disadvantages of conventional catheter approaches.
In \cite{purdy2003percutaneous}, a feasibility study of spinal subarachnoid space navigation in cadavers using a manually operated guidewire was introduced as a minimally invasive approach to the spinal cord and brain. Similarly, intraspinal navigation with catheters and fiberscopes in cadaver models was demonstrated via manual manipulation in \cite{layer2011subarachnoid}. An image-guided robot-assisted cable-driven endoscopic system for spinal interventions was presented in \cite{ascari2003new,ascari2010robot}. However, its design followed conventional principles of steerable catheters and guidewires, and did not fundamentally alter the mechanics and associated risks of tissue-device interaction. 
Kashcheev et al. demonstrated the clinical feasibility of intrathecal navigation using a $2.8\,$mm steerable flexible endoscope (thecaloscope) for visualisation and treatment of spinal arachnoid pathologies, achieving neurological improvement in $87$\% of patients; however, navigation remained entirely manual and required surgical access through laminectomy and durotomy\cite{kashcheev2017thecaloscopy}.
In summary, controlled navigation deep within confined and fragile lumens such as the spinal subarachnoid space remains elusive, motivating the need for a fundamentally different access strategy.


We pursue the alternative navigation paradigm of eversion-growing (vine-inspired) robots, introduced in \cite{hawkes2017soft} and surveyed in \cite{blumenschein2020design,al2025tip}, which depart from conventional insertion mechanics. In this approach, a robot advances by everting a tubular membrane at its distal tip, instead of sliding a body shaft through the environment. By decoupling distal navigation from proximal actuation forces, eversion-based robots minimise relative translation between the trailing body and the surrounding anatomy, reducing interaction forces while enabling follow-the-leader elongation. Early work on everting robots has demonstrated feasibility in clinical contexts such as colonoscopy \cite{borvorntanajanya2024model,shi2025design},  endovascular intervention \cite{li2021vine}, bronchoscopy \cite{xu2024novel,hwee2021everting}, and mammary duct inspection \cite{berthet2021mammobot, wu2023towards, wu2023vision}, underscoring their promise for minimally invasive procedures where reduced tissue interaction forces are critical. 

Despite the advantages of this approach, the majority of prior demonstrations have been limited to phantoms. Translation to human anatomy introduces additional constraints related to access geometry, multi-modal imaging requirements, and the mechanical compliance of living or preserved cadaveric tissue. These challenges are amplified when targeting one of the narrowest and most delicate luminal spaces in the body — the spinal subarachnoid space.

Our foundational work includes the Vine4Spine system, 
presented and validated on an up-scaled phantom in \cite{wu2025vine4spine}, and pilot human tissue studies \cite{wu2025vine}. The former established the prototypical system architecture, while the latter made clear how clinical access places additional constraints on robot deployment and navigation.  

In this work, we start with systematic optimisation and benchmarking in anatomically realistic spinal phantoms. This includes controlled evaluation of eversion-based growth, steerable navigation, endoscopic image acquisition, and quantitative interaction force assessment. Computational finite element modelling was further investigated, revealing the underlying interaction mechanism of the eversion-based growth. In phantom experiments, eversion-based navigation reduced the mean interaction force magnitude by $65.2\%$ and the peak force magnitude by $48.0\%$ compared to matched conventional push-based insertion protocols, while maintaining stable tip alignment and steerable growth over the full deployment length of $150\,$mm, including $\sim70\,$mm intrathecal navigation in the spinal subarachnoid space. Building on these findings, we conducted the first intact human cadaver demonstration of eversion-based robotic navigation within the spinal subarachnoid space under standard lumbar access, with multi-modal fluoroscopic and endoscopic guidance, and assistance from neurosurgeons. The system achieved controlled intrathecal extension and steerable advancement along the spinal canal with concurrent in-situ endoscopic imaging. No discernible tissue damage, e.g. nerve rupture, was observed when neurosurgeons post-procedurally microscopically assessed the cadaver mater.  

Together, our results establish the feasibility of eversion-based robotic growth for extended intrathecal access in real human anatomy and provide the first quantitative and procedural evidence supporting this paradigm as a mechanically viable low-friction navigation strategy for spinal subarachnoid interventions.


\begin{figure}[!t]
    \centering
    \includegraphics[width = 0.95\textwidth]{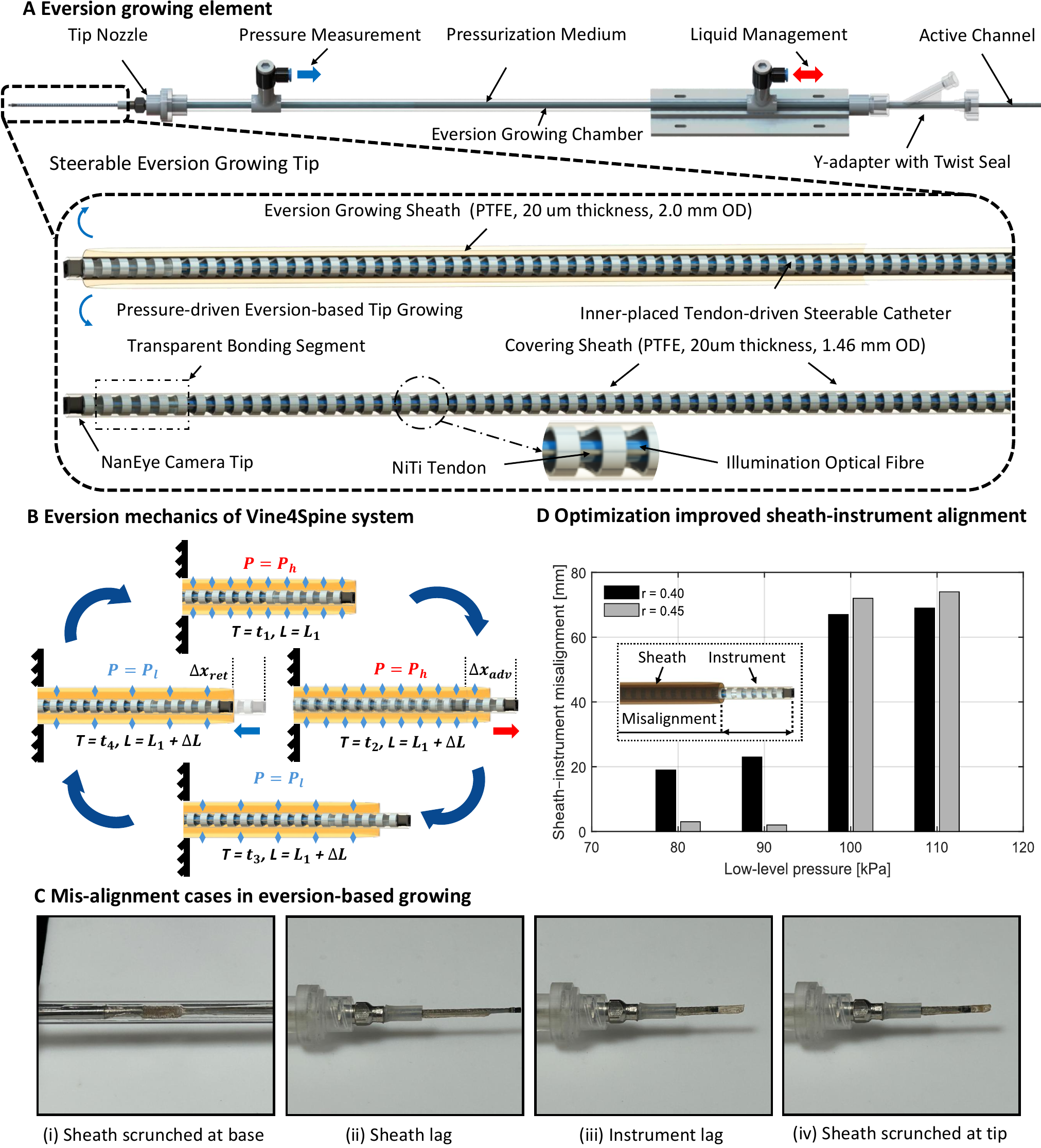}
    \caption{\textbf{Vine-inspired eversion-based growth and duty cycle benchmark:} 
    \textbf{(A)} Structural depiction of the key eversion growing elements of the Vine4Spine system. The main components include an eversion growing chamber, an eversion growing sheath, and an active channel.
    \textbf{(B)} Eversion mechanics of Vine4Spine through duty cycle strategy.
    \textbf{(C)} Effects of incomplete sheath/instrument decoupling and friction hysteresis. \textbf{(i)} Instrument lags due to its excessive retraction; \textbf{(ii)} Sheath scrunched at the base due to failed decoupling; \textbf{(iii)} Misalignment due to excessive advancement of the instrument or insufficient eversion of the sheath; \textbf{(iv)} Sheath scrunched at the tip due to excessive instrument lag. 
    \textbf{(D)} Benchmarking investigation of duty-cycle parameter optimisation. Measured sheath-instrument misalignment after the completion of full deployment of $150\,$mm with various low-level pressures and retraction ratios.} \label{fig.misalignment}
\end{figure}

\section*{Results}
\subsection*{Vine-inspired eversion-based tip growth and duty cycle mechanism}
An overview of the anatomical target environment, eversion-based robotic system, and tip-growing deployment mechanism of Vine4Spine is shown in Fig.~\ref{fig.conception}. The spinal subarachnoid space is a confined, fluid-filled, and deformable anatomical environment that complicates conventional push-based navigation [Fig.~\ref{fig.conception}A]. To address these constraints, Vine4Spine employs pressure-driven eversion growth with an inner steerable instrument with integrated imaging capabilities [Fig.~\ref{fig.conception}B]. Unlike conventional push-based catheters that generate distributed friction and tissue displacement during advancement, the eversion-based mechanism enables tip growth with minimal relative motion along the deployed body [Fig.~\ref{fig.conception}C]. 
The structural implementation of the eversion-growing element includes the folded sheath, active channel, steerable inner catheter, and embedded imaging components (\textit{Supplementary Table.~1}).
The system enables pressure-driven tip eversion, and allows proximal-to-distal access of a steerable inner-placed instrument - the only architecture that can, downstream, enable tissue sampling and therapeutic delivery.

\subsubsection*{Load-decoupled tip growth mechanism}
Unlike conventional push-based catheters, Vine4Spine advances through distal eversion of a thin-walled, compliant sheath, as depicted in Fig.~\ref{fig.misalignment}A. Growth occurs exclusively at the tip by controlled pressurisation of the folded sheath within a chamber, causing the material to evert and extend forward. Thus, the outer surface of the grown segment remains stationary relative to the surrounding tissue. 

This tip-based growth mechanism reduces relative sliding between the robot body and the intrathecal environment. In the narrow, compliant, and fluid-filled spinal subarachnoid space, where lateral clearance is only a few millimetres, minimising axial shear forces and distributed friction is critical for safe navigation.

\subsubsection*{Architecture enabling controlled eversion}

At the core of our mechanism is a pressurised chamber storing the folded eversion sheath. A low-compressibility, biocompatible liquid medium (saline-based) is used for pressurisation, enabling precise pressure regulation while reducing the risk of abrupt expansion and rupture associated with high-compressibility gases. Internal pressure is regulated in a closed-loop, based on real-time feedback, allowing growth to be continuously modulated.

The eversion growing sheath, fabricated from ultra-thin polytetrafluoroethylene (PTFE) tubing, is connected distally to a custom nozzle at the chamber outlet and proximally to the active channel. When the chamber pressure exceeds a threshold determined by sheath geometry and frictional resistance, the sheath undergoes eversion at the nozzle, turning inside out and advancing forward. Because the previously deployed material remains stationary with respect to the surrounding anatomy, axial load transmission and friction with the surrounding anatomy is inherently minimised.

An active channel serves as both a mechanical interface and a control element. It connects to the proximal end of the eversion sheath and is motorised to allow controlled translation of the sheath relative to the chamber. This enables coordinated growth and retraction, as well as precise alignment between the sheath tip and the inner-placed steerable instrument. A haemostasis valve seals the rear of the pressurisation chamber where the active channel exits, maintaining a closed fluid system.

\subsubsection*{Co-axial integration of steering and imaging}

Our configuration allows functional tools, such as miniature endoscopes or diagnostic probes, to be delivered to the distal tip without experiencing relative sliding against surrounding tissue during navigation. 

The eversion growing steerable segment of Vine4Spine comprises an inner-placed instrument that is housed within the central lumen of the eversion sheath [Fig.~\ref{fig.misalignment}A].
This inner instrument contains the patterned NiTi steerable segment that is actuated by a NiTi tendon such that catheter deflection directly dictates the trajectory of future growth. A torque-transmission shaft attached to the instrument enables rotation. Bending and translation of the inner instrument are actuated at the robot's base. The instrument further comprises a miniature endoscopic camera, an illumination optical fibre, and a covering sheath. 

The tendon and the illumination optical fibre are placed centrally within the lumen of the steerable instrument, while the cabling of the miniature endoscope is placed exteriorly and is encapsulated by the covering sheath. This coaxial architecture results in a functional instrument with a diameter of $1.45\,$mm that allows simultaneous navigation, visualisation, and future therapeutic delivery within confined intrathecal anatomy. 



\subsubsection*{Duty-cycle control for alignment stability}
Our proposed duty cycle approach utilises the internal friction between the sheath and the instrument to assist the eversion of the miniature eversion-growing sheath, which was found essential for successful and controllable eversion-based growth.
As the sheath elongates through tip eversion, the inner instrument remains stationary with respect to the external environment, thereby preserving a stable imaging and sensing viewpoint at the robot tip. Axial translation of the instrument can be actuated independently to fine-tune its protrusion beyond the sheath when required, while retraction is used to maintain alignment during growth cycles. The coaxial arrangement ensures that instrument deployment does not interfere with catheter steering or sheath eversion, enabling simultaneous navigation, visualisation, and intervention. 

While tip eversion eliminates continuous proximal pushing, maintaining alignment between the inner instrument and the compliant sheath requires coordinated pressure modulation and relative translation. To achieve stable, repeatable growth, a duty-cycle-based control strategy was implemented [Fig.~\ref{fig.misalignment}B]. During high-pressure phases ($P=P_h$), pressure-induced coupling between the eversion sheath and the inner instrument ($T=t_1$) enables controlled eversion and advancement $\Delta x_{adv}$ at the distal tip ($T=t_2$). During low-pressure phases ($P=P_l$), the coupling is reduced ($T=t_3$), allowing relative motion $\Delta x_{ret}$ between the inner instrument and the sheath through active-channel translation ($T=t_4$). This alternating sequence provides a structured framework for incremental tip growth while limiting continuous axial loading and excessive deformation of the eversion sheath. 

\subsubsection*{Translation to intrathecal deployment}

Similar to the in-situ deployment illustrated in Fig.~\ref{fig.conception}B, the Vine4Spine system is deployed into the spinal subarachnoid space via standard lumbar access and advanced under fluoroscopic guidance. Within this fluid-filled, compliant, and geometrically constrained environment, navigation is achieved through distal tip growth combined with steerable control of the inner instrument. 


The combination of liquid pressurisation, thin-walled eversion, and duty-cycle coordination provides a foundation for navigating long, narrow, and compliant anatomical spaces, i.e., the spinal subarachnoid space. However, the effectiveness of the growth strategy depends on the specific choice of duty-cycle parameters governing pressure levels and relative motions, which directly influence sheath/instrument coupling and alignment over repeated growth cycles.

\subsection*{Optimised duty-cycle parameters for aligned and stable tip growth}

Although the duty-cycle strategy enables incremental tip-based growth, its performance critically depends on the mechanics of sheath/instrument coupling during the depressurisation phase. From an idealised kinematics perspective, symmetric motion would be expected to restore alignment between the inner instrument and the eversion sheath. Given that the sheath advances at half the translational velocity of the inner instrument during the growth phase, a retraction-to-advancement ratio of $0.5$ would be expected to restore instrument tip-to-sheath tip alignment. Likewise, reducing chamber pressure to near-atmospheric levels ($P_l \sim 100\,$kPa) would theoretically minimise normal forces and decouple the inner instrument from the sheath to re-align their fronts.  
However, because sheath–instrument coupling is dominated by frictional hysteresis rather than ideal kinematics, these theoretical parameters do not translate to stable physical behaviour. Instead, persistent residual coupling was observed even at near-atmospheric pressure, indicating that nominal depressurisation leads to incomplete decoupling and non-ideal instrument/sheath sliding. 

As Fig.~\ref{fig.misalignment}C illustrates, several distinct misalignment behaviours emerged, linked to the ratio between instrument retraction and advancement $r$ and the depressurisation level $P_l$.
At near-atmospheric pressure, i.e., $P_l \sim 100\,$kPa, depressurisation was insufficient to fully decouple the inner instrument and the eversion sheath, leading to non-ideal relative sliding. The compliant sheath accumulated at its base interface with the active channel, leading to localised scrunching and significant sheath lag [Fig.~\ref{fig.misalignment}C(i,ii)]. This accumulation progressively increased friction between the sheath and the inner instrument, undermining the intended symmetry of the duty-cycle strategy. Conversely, when the retraction-to-advancement ratio was set to the theoretical value of $0.50$, the instrument tip consistently lagged behind the sheath tip [Fig.~\ref{fig.misalignment}C(iii)]. With the offset accumulating over successive cycles, the sheath gradually scrunched at the distal end of the sheath [Fig.~\ref{fig.misalignment}C(iv)], which ultimately resulted in rupture due to concentrated friction loading. These identified behaviours demonstrate that commanded symmetric motions do not translate to symmetric displacement under compliant, friction-dominated operating conditions.

Motivated by these observations, the retraction-to-advancement ratio $r$ was treated not as a geometric symmetry parameter, but as a compensation parameter to account for residual sheath/instrument coupling during the low-pressure phase. Therefore, lower ratios ($r=0.45$ and $r=0.40$) were investigated in preliminary experiments. Reducing the commanded retraction revealed that the cumulative instrument lag relative to the eversion sheath is mitigated, and incremental tip growth is consistently maintained. 
To further understand the effect of pressure on the instrument/sheath mechanical performance, low-level pressures ranging from $80\,$kPa to $110\,$kPa in steps of $10\,$kPa were evaluated in conjunction with two retraction ratios, $r=0.45$ and $r=0.40$. For each combination, the robot was deployed in free space to its full available length of $150\,$mm, and the final sheath/instrument misalignment was measured as the primary performance metric. Experiments with $r = 0.50$ were excluded from the comparison since tip scrunching consistently led to sheath rupture and failed deployment.
As Fig.~\ref{fig.misalignment}D shows, low-level pressures at or above the atmospheric level ($P_{l,ref}=100\,$kPa) resulted in significant misalignment between the sheath and the instrument, independent of the retraction ratio. Under these conditions, the sheath only covered approximately half of the instrument after its full deployment. In contrast, 
a retraction ratio of $0.45$ together with a reduced low-level pressure of $90\,$kPa demonstrated consistent alignment over the full $150\,$mm deployment. 

As a conclusion, a high-level pressure of $140\,$kPa, a low-level pressure of $90\,$kPa, and a retraction ratio of $0.45$ exhibit intended behaviour, with minimal misalignment between instrument and sheath tip even after full robot deployment. This set of parameters 
was used in all subsequent navigation experiments.

\begin{figure}[!t]
    \centering
    \includegraphics[width = 0.93\textwidth]{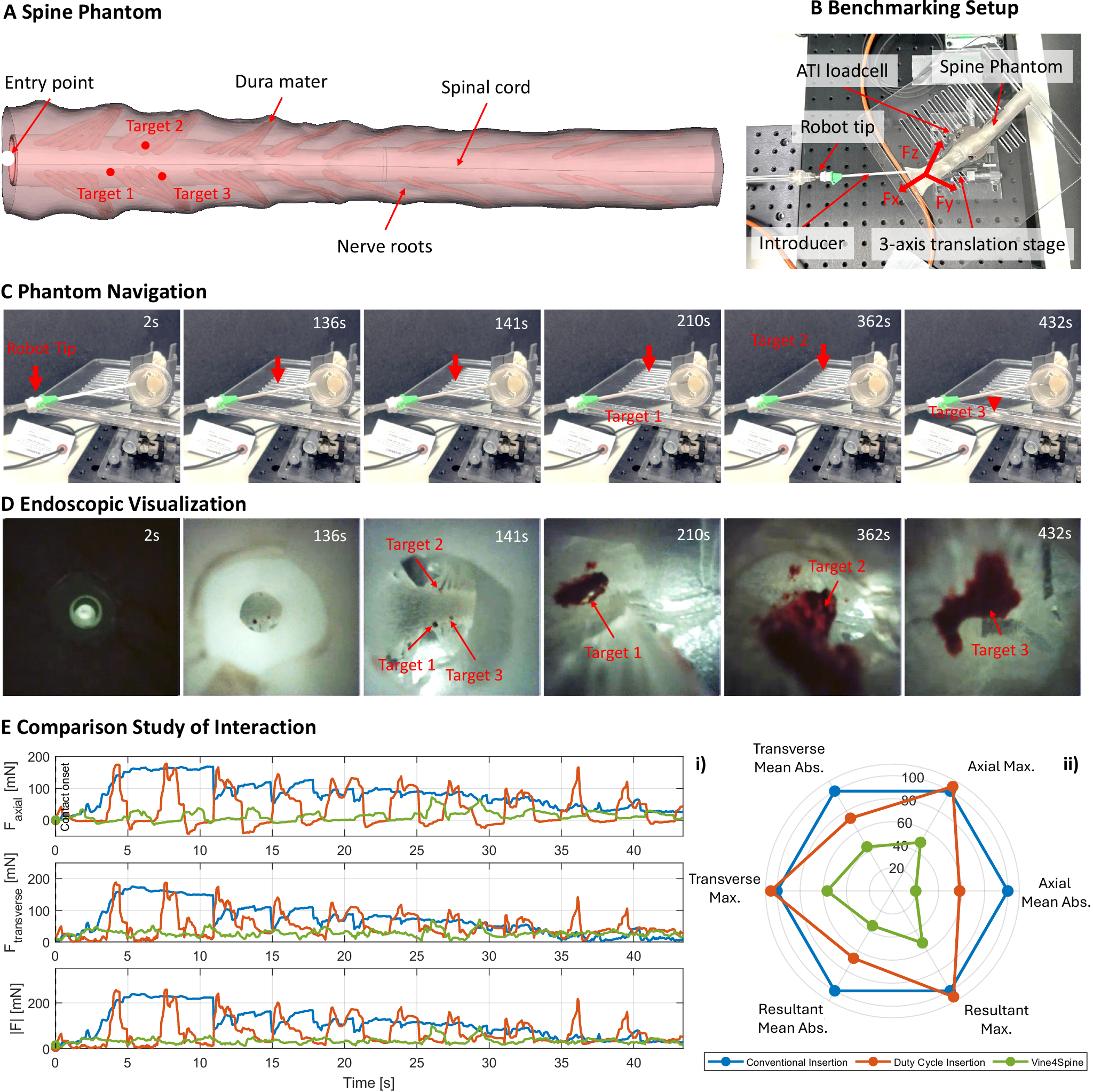}
    \caption{\textbf{Benchmarking study of Vine4Spine using a transparent spine phantom.} 
    \textbf{(A)} A cervical spine phantom is employed for the phantom study to evaluate the robot's performance, where three red markers label the navigation targets.
    \textbf{(B)} Experimental configuration of the phantom study for the evaluation of both the navigation capabilities and the reduced interaction. 
    \textbf{(C)} External phantom view when navigation is performed via Vine4Spine. \textbf{(D)} Internal view of the phantom from the endoscopic imaging. 
    \textbf{(E)} 
    Quantitative comparison of interaction forces measured during phantom navigation using \textbf{conventional catheter insertion}, \textbf{duty-cycle catheter insertion}, and the proposed \textbf{Vine4Spine system}.
    \textbf{(i)}Representative interaction force profiles illustrating the temporal evolution of axial force components $F_{axial}$, transverse force components $F_{transverse}$, and resultant force magnitude $\|\mathbf{F}\|$ during navigation. The vertical dashed line denotes the onset of tissue contact, with all trajectories temporally aligned to facilitate direct comparison of force evolution after contact.
    \textbf{(ii)} Radar plot summarizing the mean absolute and maximum values of axial, transverse, and resultant interaction forces, averaged across three independent experiments and normalized to conventional insertion. The radar plot highlights the substantial reduction in both sustained and peak interaction forces achieved by Vine4Spine compared with the two push-based navigation strategies.
    }\label{fig.phantom_exps}
\end{figure}

\subsection*{Vine4Spine enables low-force navigation in anatomically representative phantom}

Verification of growth parameters, image acquisition, steering capability, and robot/environment interaction first took place on a to-scale semi-transparent spine phantom that replicated the dura mater, central canal, and nerves (C$1$ to C$7$)\cite{chierichini2016cervical} [Fig.~\ref{fig.phantom_exps}A]. A custom introducer sheath, placed at a clinically meaningful angle for entry, enabled robot access into the phantom's subarachnoid space. The phantom was integrated onto a calibrated ATI load cell, with its X+ axis aligned with the axial direction of the spinal canal, with three markers labelled to indicate the navigation targets [Fig.~\ref{fig.phantom_exps}B]. This configuration enabled simultaneous measurement of global interaction forces ($F_x$, $F_y$, $F_z$) during teleoperated robot navigation, while providing optical access for direct observation of sheath growth and steering behaviour (\textit{Supplementary Video1,Video2}).



\subsubsection*{Mechanical and Optical Coupling for Intrathecal Endoscopic Imaging}

Imaging within the spinal subarachnoid space poses coupled mechanical and optical challenges. The environment is dark, geometrically confined, and fluid-filled, with compliant boundaries. Since it is an optically isolated environment, visualisation relies entirely on onboard illumination, which is rapidly attenuated by the scattering in the circulating fluid, light refraction, and specular reflections from wet neural surfaces. Finally, the confined and deformable geometry limits working distance and field of view (FOV). 

In such conditions, reliable visualisation depends not only on the optical arrangement, but also on the stability of the deployed robotic structure. 
Indeed, our preliminary cadaveric investigations \cite{wu2025vine} revealed that misalignment between the sheath and the imaging tip, sheath deformation, and pressure instabilities affected imaging performance beyond optical limitations.  

Our new image acquisition system addressed these challenges. The instrument incorporates a NanEye miniature chip-on-tip camera and a dedicated illumination fibre, replacing our prior borescope configuration (\textit{Supplementary Table.~2}). 
Following a series of placement experiments, we found that keeping the illumination posterior to the camera tip for $5-6\,$mm offers the best illumination effect, covering the observed FOV without saturating the camera sensor. In collaboration with neurosurgeons, this performance was optimised and validated in a mimicked spinal subarachnoid space environment, both in terms of anatomy and fluid presence.

The spine phantom was secured in a section of a pigmented sealed tube to be isolated from external illumination interference. Saline was mixed with milk and food colouring gel in red colour (Limino, US) to reproduce the absorption and scattering optical characteristics of the CSF in the cadaver. The steerable inner instrument, coupled with illumination and the chip-on-tip camera, was then inserted in the phantom. 

\begin{table}[t!]
\centering
\caption{Quantitative assessment of interaction forces using different navigation approaches}
\resizebox{0.95\textwidth}{!}{
\begin{tabular}{cccccc}
\hline
\textbf{Force Metric}         & \textbf{\begin{tabular}[c]{@{}c@{}}Conventional\\ Insertion {[}mN{]}\end{tabular}} & \textbf{\begin{tabular}[c]{@{}c@{}}Duty Cycle\\ Insertion {[}mN{]}\end{tabular}} & \textbf{\begin{tabular}[c]{@{}c@{}}Vine4Spine {[}mN{]}\end{tabular}} & \textbf{\begin{tabular}[c]{@{}c@{}}Vine4Spine reduction \\ vs Conventional\end{tabular}} & \textbf{\begin{tabular}[c]{@{}c@{}}Vine4Spine reduction\\ vs Duty Cycle\end{tabular}} \\ \hline
\textit{Axial Mean Abs.}      & $73.20\pm4.33$                                                                                   & $42.62\pm3.04$                                                                                 & $14.76+2.51$                                                                       & $\downarrow79.8\%$                                                                                   & $\downarrow65.4\%$                                                                                \\
\textit{Axial Max.}           & $172.87\pm7.10$                                                                                  & $180.72\pm3.09$                                                                                & $84.13\pm15.58$                                                                      & $\downarrow51.3\%$                                                                                   & $\downarrow53.4\%$                                                                                \\
\textit{Transverse Mean Abs.} & $66.84\pm3.47$                                                                                   & $48.69\pm3.37$                                                                                 & $29.53\pm13.74$                                                                      & $\downarrow55.8\%$                                                                                   & $\downarrow39.3\%$                                                                                \\
\textit{Transverse Max.}      & $176.63\pm10.16$                                                                                 & $186.09\pm2.31$                                                                               & $99.89\pm25.70$                                                                      & $\downarrow43.4\%$                                                                                   & $\downarrow46.3\%$                                                                                \\
\textit{Resultant Mean Abs.}  & $100.12\pm5.54$                                                                                  & $67.44\pm4.30$                                                                                 & $34.83\pm12.80$                                                                      & $\downarrow65.2\%$                                                                                   & $\downarrow48.4\%$                                                                                \\
\textit{Resultant Max.}       & $243.09\pm14.08$                                                                                 & $257.68\pm2.24$                                                                                & $126.39\pm31.80$                                                                     & $\downarrow48.0\%$                                                                                 & $\downarrow50.9\%$                                                                                \\ \hline
\end{tabular}\label{table.force}
}
\end{table}

\subsubsection*{Controlled Steering and Targeting in Confined Anatomy}

Further experiments assessed whether tip-based eversion and tendon-driven steering could achieve controlled navigation within the confined lumen of the phantom. After extending past the introducer sheath, the steerable tip was teleoperated toward three predefined neural targets marked within the canal.

As illustrated in Fig.~\ref{fig.phantom_exps}C and D, the robot sequentially navigated toward the three labelled targets, with both internal endoscopic imaging and external video confirmation of trajectory control. Growth occurred through incremental eversion without observable buckling or proximal body translation. The ability to steer after incremental extension demonstrates that distal curvature modulation is preserved during eversion-based growth, even within a constrained millimetre-scale environment.

While these steering experiments confirm controllability, the primary objective of the phantom study was to quantify interaction forces relative to conventional insertion strategies, as described next.

\subsubsection*{Quantitative Assessment of Interaction Forces}
To quantify tissue-robot interaction during navigation, we carried out a comparative study among three navigation strategies: \textbf{(1)} continuous \textbf{conventional} catheter \textbf{insertion}, \textbf{(2)} \textbf{duty-cycle} controlled catheter \textbf{insertion}, and \textbf{(3)} the proposed eversion-based \textbf{Vine4Spine} system. 
The duty-cycle protocol was designed to match the temporal advancement sequence, cumulative insertion distance and average translational velocity of Vine4Spine, thereby isolating the influence of the eversion-based growth mechanism. Global force measurements
($F_x$, $F_y$, $F_z$) were resolved into the axial force component $F_{axial}$, transverse force component $F_{transverse}$, and resultant force magnitude $\|\mathbf{F}\|$ for comparison (\textit{Supplementary Table.~3, Fig.~1, Fig.~2, and Fig.~3}). 

The three navigation strategies exhibited fundamentally different interaction-force characteristics throughout phantom navigation [Fig.~\ref{fig.phantom_exps}E]. Representative force profiles revealed distinct mechanical loading behaviours immediately after tissue contact [Fig.~\ref{fig.phantom_exps}E(i)]. Continuous catheter insertion produced a rapid increase in interaction force, followed by a prolonged period of elevated loading throughout advancement. The interaction force remained consistently high over much of the insertion trajectory, indicating persistent mechanical contact between the catheter shaft and the surrounding anatomy.

Duty-cycle insertion altered this behaviour by periodically interrupting continuous advancement. Instead of a sustained force plateau, the interaction profile consisted of repeated loading and unloading cycles, resulting in lower average interaction forces than continuous insertion. Nevertheless, each advancement phase generated pronounced transient force peaks, and the peak interaction forces remained comparable to those observed during conventional insertion.

In contrast, Vine4Spine exhibited a markedly different interaction pattern. Throughout navigation, interaction forces remained consistently low with substantially fewer high-amplitude fluctuations, producing the smoothest force profile among the three approaches. This behaviour is consistent with the localised tip-growth mechanism of Vine4Spine, which minimises sliding of the deployed body against surrounding tissues and therefore limits sustained shaft–tissue interaction. The radar plot [Fig.~\ref{fig.phantom_exps}E(ii)] further summarises these trends by comparing the normalised mean absolute and maximum axial, transverse, and resultant interaction forces across the three navigation strategies. Compared with both push-based approaches, Vine4Spine consistently occupied the smallest radar area, illustrating a simultaneous reduction in sustained and peak interaction forces across all evaluated force components.

Quantitative analysis across three independent experiments further confirmed these observations [Table~\ref{table.force}]. Vine4Spine consistently reduced both sustained and peak interaction forces relative to the two push-based approaches. Compared with conventional insertion, the mean absolute axial, transverse, and resultant interaction forces were reduced by $79.8\%$, $55.8\%$, and $65.2\%$, respectively, while the corresponding peak forces decreased by $51.3\%$, $43.4\%$, and $48.0\%$. Relative to duty-cycle insertion, Vine4Spine achieved further reductions of $65.4\%$, $39.3\%$, and $48.4\%$ in the mean absolute axial, transverse, and resultant forces, together with reductions of $53.4\%$, $46.3\%$, and $50.9\%$ in the corresponding peak forces. Although duty-cycle insertion reduced sustained loading compared with continuous insertion, it did not substantially mitigate peak interaction forces because each advancement cycle still relied on proximal pushing of the catheter. By contrast, Vine4Spine reduced both sustained and transient interaction forces, demonstrating a fundamentally different tissue-interaction mechanism.

Importantly, reductions were consistently observed in both the axial and lateral force components, indicating that the reduced interaction was not confined to the insertion direction but extended across the overall tissue–robot contact interface. Collectively, these phantom experiments demonstrate that eversion-based growth fundamentally alters the interaction profile during navigation, suppressing both sustained loading and high transient force events while maintaining the lowest overall interaction forces among the three navigation strategies.

All force benchmarking experiments were performed over the maximum available deployment length of $150\,$mm, with force analysis beginning once the robot exited the introducer sheath and entered the spinal subarachnoid space. Throughout the evaluated range, the optimised duty-cycle parameters maintained stable sheath–instrument alignment without cumulative scrunching or structural failure.

\begin{figure}[!t]
    \centering
    \includegraphics[width=0.76\textwidth]{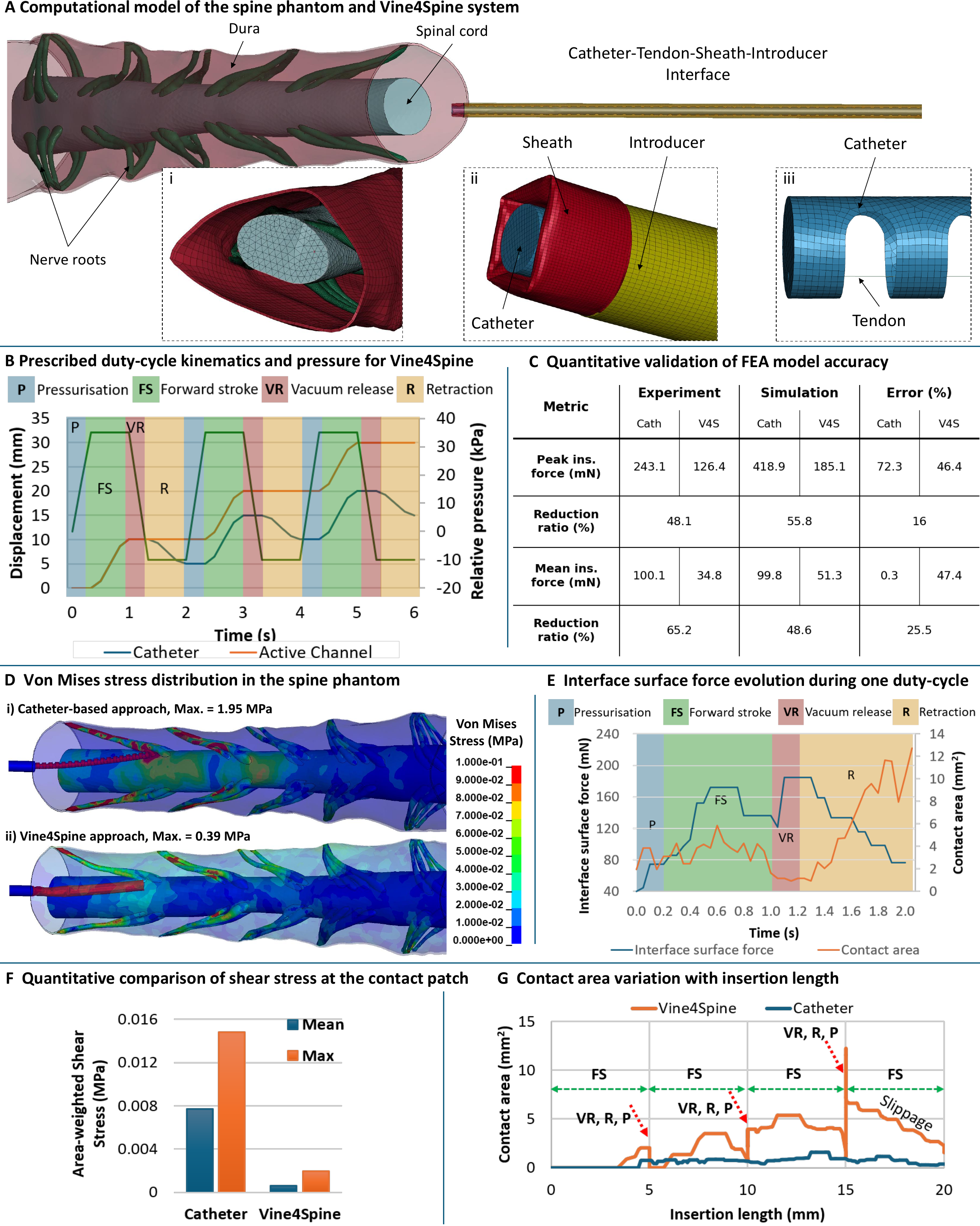}
    \caption{\textbf{Finite element model and biomechanical analysis of robot–tissue interaction.} 
    \textbf{(A)} Overview of the finite element model, comprising the spinal phantom (dura mater, spinal cord, and nerve roots), steerable catheter, tendon, PTFE everting sheath, and rigid introducer, with enlarged views of \textbf{(i)} the phantom geometry, \textbf{(ii)} the catheter-sheath-introducer assembly, and \textbf{(iii)} the catheter-tendon finite element meshes; 
    \textbf{(B)} Prescribed duty-cycle kinematics over the first three insertion cycles, showing catheter and active-channel displacements (left axis) together with the corresponding internal pressure profile (right axis), with the four phases indicated: \textbf{(1)} pressurisation (P), \textbf{(2)} forward stroke (FS), \textbf{(3)} vacuum release (VR), \textbf{(4)} retraction (R); 
    \textbf{(C)} Comparison of experimentally measured and simulated mean and peak insertion forces for conventional catheter insertion (Cath) and Vine4Spine (V4S) to evaluate the accuracy of FEA modelling.
    \textbf{(D)} Contours of von Mises stress (MPa) distribution on the phantom at a representative insertion phase for naked-catheter (top) and Vine4Spine (bottom) insertion; 
    \textbf{(E)} Interface surface force over one representative duty-cycle period, annotated with the four insertion phases; 
    \textbf{(F)} Mean and maximum area-weighted interfacial shear stress (MPa) over the active contact patch for both insertion modalities; 
    \textbf{(G)} Contact area between robot and phantom as a function of the insertion length for naked catheter and Vine4Spine insertion, showing how the contact area evolves during the forward stroke and how it is affected by the other three duty-cycle phases.}
    \label{fig:combined_results}
\end{figure}

\subsection*{Computational modelling reveals the biomechanical mechanisms underlying reduced tissue interactions}

To further investigate biomechanical quantities that could not be directly measured experimentally, including distributed interfacial stresses and contact mechanics within the spinal subarachnoid space, we developed a nonlinear finite element model replicating the experimental phantom and robotic system with duty cycle implementation [Fig.~\ref{fig:combined_results}A, B].
The model incorporated the catheter, actuation tendon, eversion sheath, and introducer, along with a representative phantom composed of dura mater, spinal cord, and nerve roots, thereby providing time-resolved interfacial information during their interactions. This computational framework replicated the experimental insertion protocols for both conventional
push-based insertion and Vine4Spine navigation, enabling quantitative comparison between simulated and experimentally measured interaction forces while providing spatially resolved mechanical information inaccessible through experimental measurements.

\subsubsection*{Physics-based computational model presents consistent agreement with experimental observations}

 The computational model reproduced the experimentally observed reduction in interaction forces achieved by Vine4Spine with reasonable fidelity [Fig.~\ref{fig:combined_results}C]. The primary validation metric, i.e., the force reduction ratio between Vine4Spine and conventional catheter insertion, exhibited close agreement between simulation and experiment: the simulated
peak-force reduction ratio of \SI{55.8}{\percent} differed from the
experimentally measured value of \SI{48.1}{\percent} by only
\SI{16}{\percent}, while the simulated mean-force reduction ratio of
\SI{48.6}{\percent} differed from the experimental value of
\SI{65.2}{\percent} by \SI{25.5}{\percent}. Consistent directional agreement in force reduction was maintained across both metrics, supporting the use of the model as a mechanistic investigation
tool for tissue-level biomechanical quantities inaccessible through
experimental measurement alone.
 
Regarding absolute force magnitudes, agreement was metric-dependent. For conventional catheter insertion, the simulated mean interaction force (\SI{99.8}{\milli\newton}) was in excellent agreement with the experimental value (\SI{100.1}{\milli\newton}), yielding a discrepancy of only \SI{0.3}{\percent}. However, the simulated peak insertion force (\SI{418.9}{\milli\newton}) substantially overestimated the experimental value (\SI{243.1}{\milli\newton}) by \SI{72.3}{\percent}. For Vine4Spine,
both the simulated peak (\SI{185.1}{\milli\newton}) and mean
(\SI{51.3}{\milli\newton}) insertion forces overestimated the corresponding experimental values (\SI{126.4}{\milli\newton} and \SI{34.8}{\milli\newton}) by \SI{46.4}{\percent} and \SI{47.4}{\percent}, respectively. These overestimations are primarily attributable to two sources of discrepancy: (i) the inherent difference between the simulated interfacial contact force,
computed directly at the device-phantom interface, and the experimentally measured load cell signal, which captures the reaction force transmitted through the phantom to an external fixture and might involve compliance and partial force cancellation along the transmission path; (ii) the uncertainty in the friction coefficients assigned to the device-phantom contact interfaces, which were derived from tribological data for proxy materials in the absence of direct Agilus30 measurements, disproportionately affecting peak force predictions that are sensitive to the assumed static friction coefficient at the onset of sliding. 
Despite these absolute discrepancies, the simulation consistently reproduced the qualitative force evolution across the duty-cycle phases [Fig.~\ref{fig:combined_results}E] and the correct ordering of interaction forces between the two modalities, confirming its suitability for mechanistic rather than predictive force estimation. 

Overall, the consistent directional agreement in force reduction across both peak and mean metrics, and across both simulation and experiment, supports the validity of the computational framework for investigating tissue-level biomechanical interactions beyond experimental measurements.

\subsubsection*{Eversion mechanism redistributes tissue loading and reduces local stress}

Having established agreement with the experimental force measurements, we next examined the spatial distribution of tissue loading during navigation.
 
The contours of the von Mises stress distribution in the phantom at a representative insertion depth with the tip touching a nerve root reveal a marked difference in load distribution between the two insertion modalities [Fig.~\ref{fig:combined_results}D]. Navigating through a conventional
push-based catheter, stress was concentrated primarily around the advancing
tip and the adjacent nerve root attachments, reaching a peak value of
\SI{1.95}{\mega\pascal}. Instead, for Vine4Spine, the stress field was
visibly more uniformly distributed along the everted sheath length, with a
substantially lower peak value of \SI{0.39}{\mega\pascal}. Nonetheless,
localised elevated stresses remained at certain nerve root regions; however,
Vine4Spine reduced the peak von Mises stress by approximately five-fold,
indicating substantially more distributed tissue loading.
 
Analysis of the interface force and contact area throughout an individual
duty-cycle period further illustrated the dynamic loading behaviour of
Vine4Spine [Fig.~\ref{fig:combined_results}E]. Interaction forces increased
during sheath pressurisation (P) and forward advancement (FS), with a further slight increase during vacuum-assisted release (VR), before
progressively decreasing during catheter
retraction (R), consistent with the four-phase insertion strategy
implemented experimentally. Simultaneously, the contact area between the
device (sheath) and the phantom wall evolved in synchrony with the duty-cycle phases. In particular, it expanded slowly during forward strokes, where the sheath was under pressure, it partially contracted during VR, where the sheath tip was flash with the catheter tip, and expanded rapidly during R, where the catheter retracted within the sheath, leaving a vacuumed portion of the sheath laying flat on the phantom wall. This
reflects the continuous redistribution of tissue contact across the
everted sheath. This cyclic loading profile closely mirrors the
interaction-force evolution observed during the phantom experiments and
further supports the proposed eversion-based mechanism of distributed tissue
loading rather than localised force accumulation through continuous pushing.
 
Corroborating the trends observed in the von Mises stress field, quantitative
analysis of the area-weighted interfacial shear stress over the active
contact region revealed markedly different load distributions between the
two insertion strategies [Fig.~\ref{fig:combined_results}F]. The reduction in local tissue
loading was clearly supported by both metrics: Vine4Spine reduced the mean
area-weighted interfacial shear stress by approximately \SI{92}{\percent} and the peak
shear stress by approximately \SI{86}{\percent}, compared to
conventional catheter insertion.

\begin{figure}[!t]
    \centering
    \includegraphics[width = 0.75\textwidth]{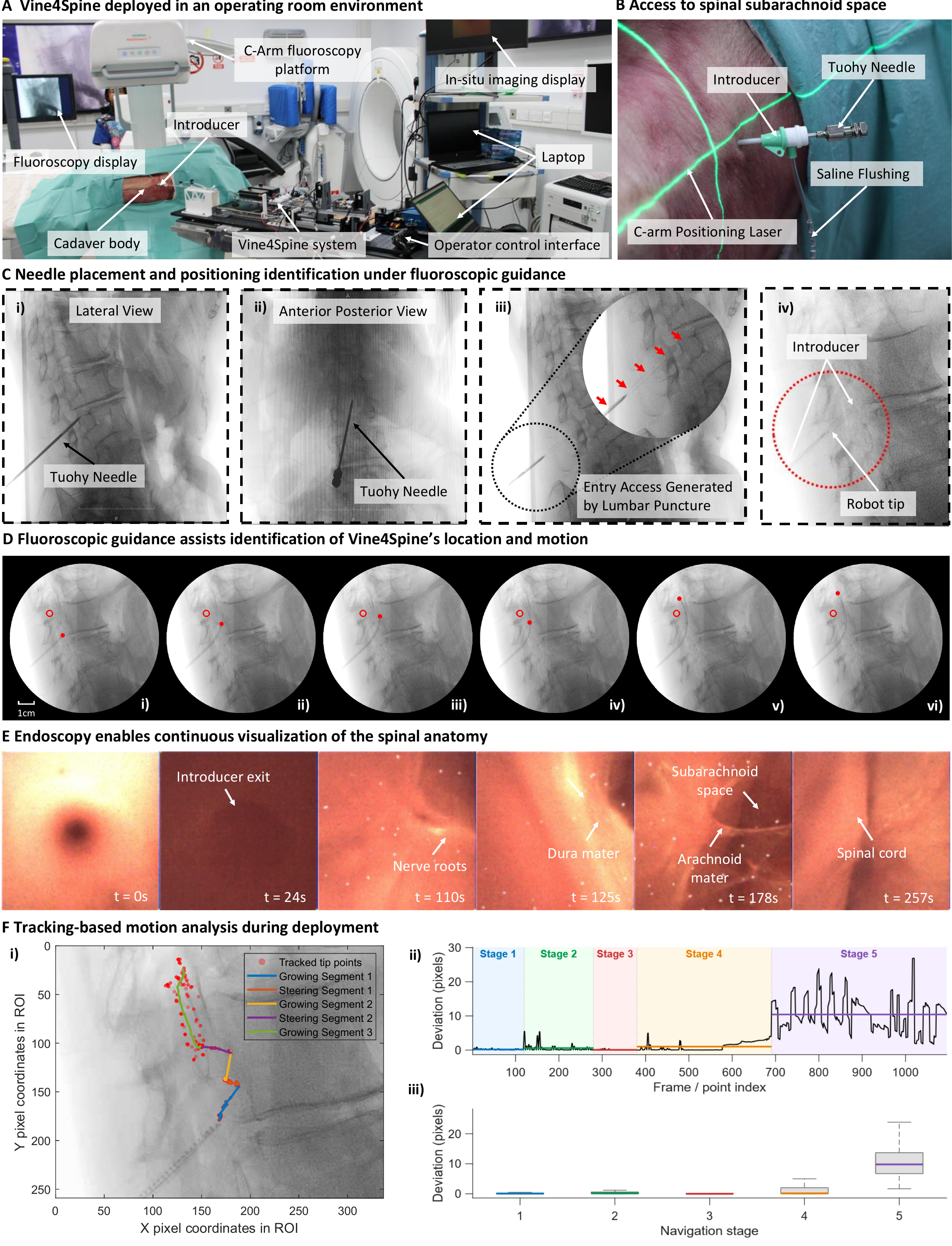}
    \caption{\textbf{Human cadaveric validation of Vine4Spine during intrathecal navigation.}
    \textbf{(A)} Experimental deployment of the Vine4Spine system in a mock operating-room environment for navigation within the spinal subarachnoid space of an intact human cadaver. 
    \textbf{(B)}Customised access interface used to establish intrathecal entry via lumbar puncture while integrating the introducer, Tuohy needle, and saline flushing system.
    \textbf{(C)} Fluoroscopy-guided generation and verification of the intrathecal access, including needle placement in lateral and anteroposterior views \textbf{(i-ii)}, identification of the lumbar puncture entry site \textbf{(iii)}, and confirmation of robot deployment through the introducer \textbf{(iv)}.
    \textbf{(D)} Representative fluoroscopic sequence illustrating the advancement of Vine4Spine within the spinal subarachnoid space during navigation.
    \textbf{(E)} Representative in situ endoscopic images acquired during deployment, demonstrating continuous visualisation of intrathecal anatomical structures, including the introducer exit, nerve roots, dura mater, arachnoid mater, subarachnoid space, and spinal cord.
    \textbf{(F)} Motion analysis of robot deployment based on fluoroscopic tracking.
    \textbf{(i)} Reconstructed tip trajectory segmented according to the different deployment stages.
    \textbf{(ii)} Point-wise deviation between the tracked and filtered trajectories across the deployment stages.
    \textbf{(iii)} Distribution of trajectory deviation for each deployment stage, demonstrating stage-dependent deployment behaviour.
   }\label{fig.mockor}
\end{figure}

\subsubsection*{Distributed contact mechanics explain the reduction in
interaction forces}
 
Analysis of the contact area variation with insertion length further
illustrated the mechanistic difference between the two navigation strategies [Fig.~\ref{fig:combined_results}G]. Whereas the conventional catheter maintained only a small, localised contact region throughout insertion, Vine4Spine progressively enlarged the tissue-device interface, reaching a maximum contact area of approximately \SI{12.5}{\milli\meter\squared} during advancement. Contact area increased during each forward advancement (FA) phase and partially reduced during the vacuum release (VR), retraction (R),
and pressurisation (P) phases, reflecting the cyclic eversion kinematics.
A transient slippage event was observed at approximately
\SI{15}{\milli\meter} insertion, corresponding to a mechanical decoupling between the everted sheath and the phantom wall at deeper insertion depths, after which the contact area partially recovered as the sheath re-engaged the phantom.
 
The expanded contact interface distributes interaction forces over a
substantially larger tissue surface, thereby reducing local traction and mitigating the stress concentrations generated by conventional push-based insertion. This cyclic loading profile closely mirrors the interaction-force evolution observed during the phantom experiments and further supports the proposed eversion-based mechanism of
distributed tissue loading rather than force accumulation through continuous pushing.
 
Collectively, these computational analyses demonstrate that eversion-based growth fundamentally changes the mechanics of tissue interaction by replacing concentrated tip loading with distributed contact along the deployed sheath. This mechanistic explanation complements the experimentally observed reductions in interaction force and provides a biomechanical rationale for the subsequent cadaveric validation of Vine4Spine.

\subsection*{Vine4Spine demonstrates safe navigation in a human cadaveric spine}

\subsubsection*{Cadaveric deployment and access feasibility}

To evaluate the feasibility of eversion-based robotic growth in human anatomy, Vine4Spine was deployed within the spinal subarachnoid space of an intact fresh-frozen human cadaver (male, $92$ years old, no known spinal pathology) under neurosurgical supervision in a mock operating-room environment [Fig.~\ref{fig.mockor}]. Following thawing to restore soft-tissue mechanical properties, intrathecal access was established through a standard fluoroscopy-guided lumbar puncture using a conventional Tuohy needle and introducer, without surgical exposure of the spinal canal [Fig.~\ref{fig.mockor}A-C]. The robotic platform was positioned adjacent to the operating table without obstructing the surgical workspace, while fluoroscopy-assisted needle placement identified the optimal lumbar puncture trajectory before robot deployment.




To facilitate fluoroscopic visualisation, the pressure-regulation fluid was mixed with a radiographic contrast agent. A clinical introducer equipped with a sealed irrigation port provided a lubricated access pathway, reducing insertion resistance while maintaining a fluid-filled environment consistent with cerebrospinal fluid and improving endoscopic visualisation during navigation. Through this conventional clinical access route, Vine4Spine was deployed using pressure-driven distal eversion while steering was achieved by the inner catheter. To the best of our knowledge, this represents the first demonstration of controlled eversion-based robotic growth and steering within the intact human spinal subarachnoid space.

\subsubsection*{Controlled intrathecal deployment via load-decoupled tip growth}

Prior to intrathecal deployment, approximately $50\,$mm of eversion length was preloaded into the introducer, while continuous saline flushing maintained a lubricated fluid interface at the access site to minimise insertion resistance and prevent air ingress [Fig.~\ref{fig.mockor}B]. 
Following fluoroscopic confirmation of successful intrathecal access [Fig.~\ref{fig.mockor}C], Vine4Spine was deployed using the pressure-regulated duty-cycle strategy described above (\textit{Supplementary Video3}). 

Under this control regime, the robot achieved continuous load-decoupled intrathecal tip growth over its full available extension of $150\,$mm. , while maintaining coaxial alignment between the steerable catheter and the everting sheath. Throughout deployment, multimodal imaging combining fluoroscopic guidance and the embedded chip-on-tip endoscope confirmed stable eversion without buckling, sheath collapse, or loss of alignment [Fig.~\ref{fig.mockor}D,E]. The fluoroscopic sequence further demonstrated progressive advancement of the robot within the spinal subarachnoid space, while the onboard endoscope provided continuous visualisation of key intrathecal anatomical structures, including the dura mater, arachnoid mater, nerve roots and spinal cord (\textit{Supplementary Video4, Video5}).

\begin{figure}[!t]
    \centering
    \includegraphics[width =0.8\textwidth]{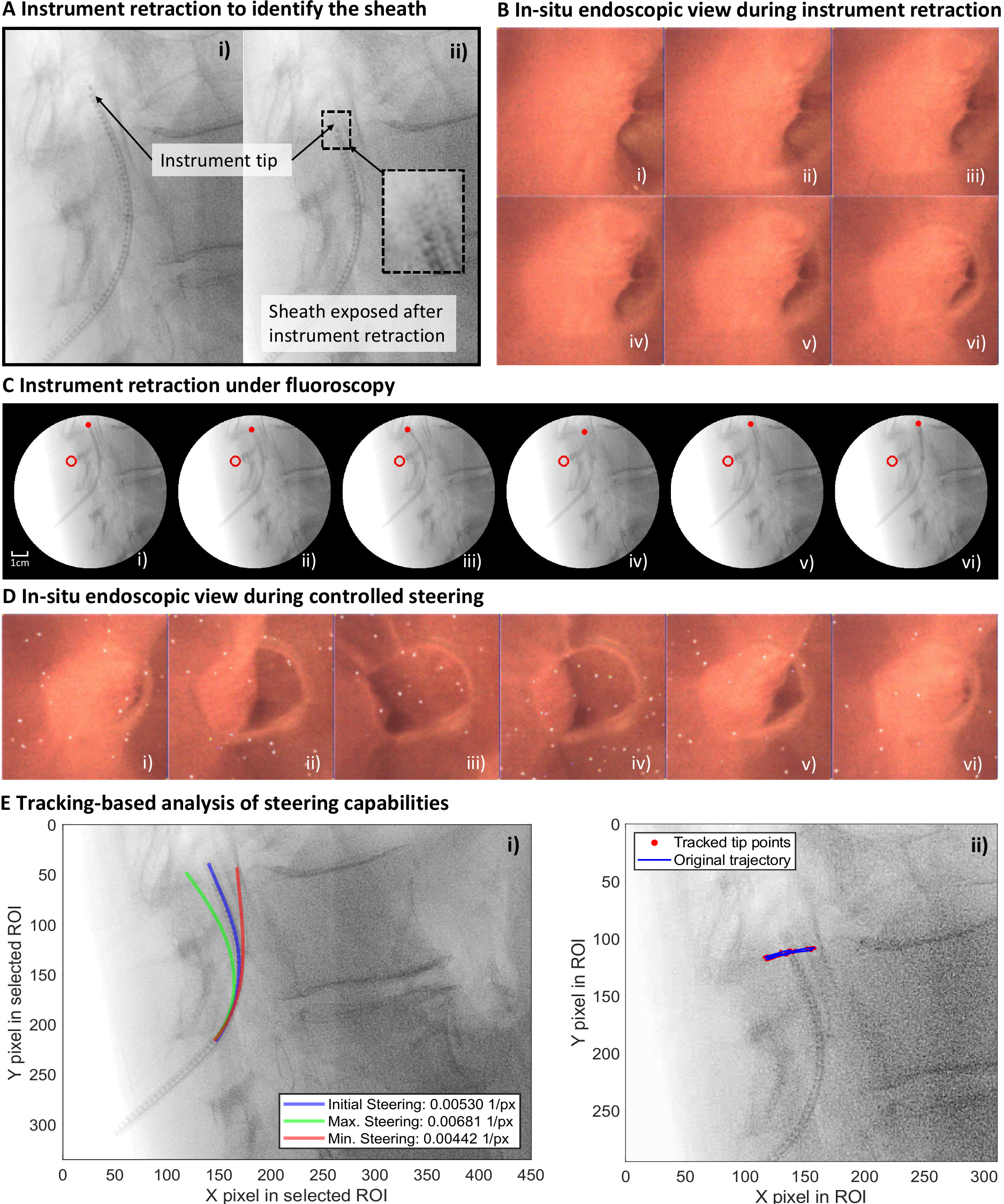}
    \caption{\textbf{Functional validation of Vine4Spine after full intrathecal deployment.} 
    \textbf{(A)} Fluoroscopic assessment of the instrument–sheath configuration following full deployment. 
    \textbf{(i)} Fully deployed Vine4Spine with the steerable catheter positioned inside the everted sheath. 
    \textbf{(ii)} Retraction of the inner catheter exposes the stationary everting sheath, confirming preservation of the deployed sheath geometry and the coaxial instrument–sheath alignment.
    \textbf{(B)} Sequential in situ endoscopic images acquired during catheter retraction, demonstrating progressive exposure of the inner surface of the deployed sheath while maintaining a sealed and stable deployment configuration.
    \textbf{(C)} Representative fluoroscopic sequence during controlled steering of the deployed robot. A red circular marker highlights catheter motion, while a red dot indicates the catheter tip.
    \textbf{(D)} Representative endoscopic views acquired during steering, illustrating simultaneous visualisation of the surrounding spinal anatomy and the inner contour of the deployed everting sheath throughout steering.
    \textbf{(E)} Quantitative evaluation of steering behaviour based on fluoroscopic tracking.
    \textbf{(i)} Reconstruction of the instrument centrelines at the initial, maximum, and minimum steering configurations, quantifying the steering workspace following full deployment.
    \textbf{(ii)} Semi-automatic tracking of the distal instrument tip within a manually selected region of interest using optical flow and Kalman filtering, demonstrating an approximately $\sim50$-pixel ($\sim12$-mm) steering displacement.
    }\label{fig.retract4view}
\end{figure}


\subsubsection*{Tracking-based motion analysis demonstrates stage-dependent deployment behaviour}
Tracking of the robot tip reconstructed the complete intrathecal navigation trajectory, revealing three sequential growth phases interleaved with two steering adjustments [Fig.~\ref{fig.mockor}F(i)].  Throughout navigation, distal advancement was maintained by localised tip eversion while preserving a stable trajectory within the spinal subarachnoid space (\textit{Supplementary Video6}).

To quantify deployment dynamics, the deviation between the tracked tip trajectory and its filtered representation was analysed throughout navigation [Fig.~\ref{fig.mockor}F(ii,iii)]. Rather than reflecting tracking uncertainty, this deviation represents the oscillatory component introduced by periodic duty-cycle actuation and therefore provides a quantitative measure of the coupled robot–environment dynamics. The measured response reflects the combined influence of cyclic actuation, increasing deployed length, accumulated steering curvature, and interaction with the compliant spinal anatomy.

Minimal deviations were observed during Stages 1–3, indicating smooth tip progression with limited oscillatory behaviour during the initial growth and steering phases. Motion variability increased modestly during Stage 4 and became more pronounced during the final deployment stage (Stage 5), as shown by both the temporal deviation profile and the broader distribution of deviation values in the stage-wise boxplot. The increased oscillatory behaviour coincided with progressive robot deployment, accumulated steering curvature, and continued interaction with the confined and compliant spinal subarachnoid space, suggesting an increasing influence of both robot configuration and environmental constraints on the dynamic response of the system.

The stage-dependent evolution of the deviation demonstrates distinct dynamic characteristics associated with successive deployment phases while confirming predictable transitions between growth and steering. Importantly, despite the increase in oscillatory motion during the later stages, the robot maintained continuous distal tip advancement without interruption throughout the complete deployment sequence, indicating that the periodic oscillations induced by duty-cycle actuation and robot-environment interactions did not compromise deployment stability or controllability.


Taken together, these results demonstrate that Vine4Spine achieves smooth, controlled, and robust load-decoupled tip growth throughout intrathecal navigation, highlighting the robustness of the proposed multi-stage deployment strategy for navigating the confined and compliant spinal subarachnoid space.

\subsubsection*{Instrument–sheath alignment and sealing integrity preserved throughout deployment}

To determine whether pressure-driven eversion preserved the intended robot architecture during intrathecal navigation, fluoroscopic imaging was combined with in situ endoscopic visualisation after full deployment [Fig.~\ref{fig.retract4view}A,B]. Retraction of the inner instrument progressively exposed the everted sheath under fluoroscopic guidance, while the embedded miniature endoscope simultaneously visualised the sheath boundary from within the robot. The controlled and progressive appearance of the sheath during instrument retraction demonstrated that coaxial alignment between the inner instrument and the deployed everting sheath was maintained throughout navigation.

The endoscopic images showed an unobstructed imaging field within the exposed sheath boundary, whereas regions outside the sheath contour appeared increasingly blurred because of optical interference from the semi-transparent sheath material. Corresponding fluoroscopic views confirmed controlled sheath exposure without observable displacement, collapse, or structural distortion, further supporting preservation of the instrument–sheath configuration after deployment. Throughout navigation and repeated duty-cycle actuation, no interruption of eversion, loss of chamber pressure, or fluoroscopic evidence of contrast leakage was observed. The absence of contrast dispersion indicated that the pressurised chamber remained sealed, demonstrating that the eversion architecture tolerated repeated pressure cycling within the confined and compliant spinal subarachnoid space without compromising structural integrity.

\subsubsection*{Distal steering authority is preserved following extended intrathecal deployment}

Following full intrathecal deployment, the steering capability of Vine4Spine was evaluated under simultaneous fluoroscopic and endoscopic guidance [Fig.~\ref{fig.retract4view}C-E]. Partial retraction of the inner instrument exposed the contrast-filled distal sheath for fluoroscopic visualisation, after which tendon actuation generated controlled bending while preserving the deployed eversion structure. Fluoroscopy enabled continuous localisation of the distal instrument relative to the surrounding anatomy, whereas the embedded endoscope provided direct visualisation of the corresponding changes within the cerebrospinal fluid corridor. Representative fluoroscopic and endoscopic sequences confirmed distal steering without observable buckling, instrument–sheath misalignment, sheath displacement, or collapse [Fig.~\ref{fig.retract4view}C,D]  (\textit{Supplementary Video7,8}).

Tracking-based analysis quantified both the change in bending curvature and the resulting distal-tip motion [Fig.~\ref{fig.retract4view}E] (\textit{Supplementary Video9}). Reconstruction of the instrument centreline under the initial, maximum, and minimum steering configurations demonstrated reproducible modulation of the deployed curvature, with image-plane curvatures of $5.30e^{-3}$, $6.81e^{-3}$, and $4.42e^{-3}$ $pixel^{-1}$, respectively [Fig.~\ref{fig.retract4view}E(i)]. These measurements confirm that tendon actuation remained capable of generating controlled bidirectional curvature changes despite the fully deployed eversion sheath, presenting reproducible modulation of the distal curvature across the achievable steering workspace. Semi-automatic tip tracking further measured a distal-tip displacement of approximately $50$ pixels, corresponding to approximately $12\,$mm during teleoperated steering [Fig.~\ref{fig.retract4view}E(ii)], confirming effective transmission of steering motion through the deployed instrument–sheath assembly. Representative fluoroscopic and endoscopic sequences further verified that distal bending was achieved without observable buckling, instrument–sheath misalignment, sheath displacement, or collapse of the eversion sheath.

Steering responsiveness was retained after intrathecal extension of up to $150\,$mm, including approximately $80\,$mm of deployment beyond the introducer and into the spinal subarachnoid space. Thus, incremental load-decoupled growth did not compromise distal curvature control despite the increasing deployed length. Together, these findings demonstrate that Vine4Spine preserves steering authority, coaxial alignment, chamber sealing, and structural integrity following clinically relevant intrathecal deployment.

\subsubsection*{Multimodal imaging confirms mechanical and optical integrity}

Fluoroscopic and in situ endoscopic imaging provided complementary external and internal validation of the deployed system. Steering adjustments observed fluoroscopically produced concordant changes in the endoscopic field of view, confirming that distal curvature modulation translated into corresponding changes in anatomical orientation. Likewise, fluoroscopically observed advancement was accompanied by progressive motion of anatomical features within the endoscopic view. This multimodal correspondence supports the accuracy of the observed deployment and steering behaviour while confirming maintenance of imaging continuity throughout navigation.

\subsubsection*{Qualitative safety observations}
No gross disruption of the dura mater or surrounding neural structures was observed under real-time fluoroscopic or in-situ imaging. The subarachnoid space remained sealed throughout the process, indicating that the dura remained intact. The robot maintained central alignment along the longitudinal axis of the spinal canal, without evidence of tissue dragging, focal compression, or abrupt trajectory deviation. No uncontrolled sheath deformation occurred during eversion or retraction phases. No leakage of the pressurisation medium was observed, even when $P_h$ was artificially increased to $190\,$kPa (close to the burst pressure of the sheath) during the preoperative testing of sealing, well above the nominal operating pressure; this indicates a safe operating margin. Also, the system was transiently pressurised to $180\,$kPa, the maximum working pressure, under closed-loop regulation during the cadaver study. No fluoroscopic evidence of contrast leakage was detected during the entire operational procedure, confirming structural robustness and sealing integrity during robotic navigation.

\begin{figure}[!t]
    \centering
    \includegraphics[width = 0.95\textwidth]{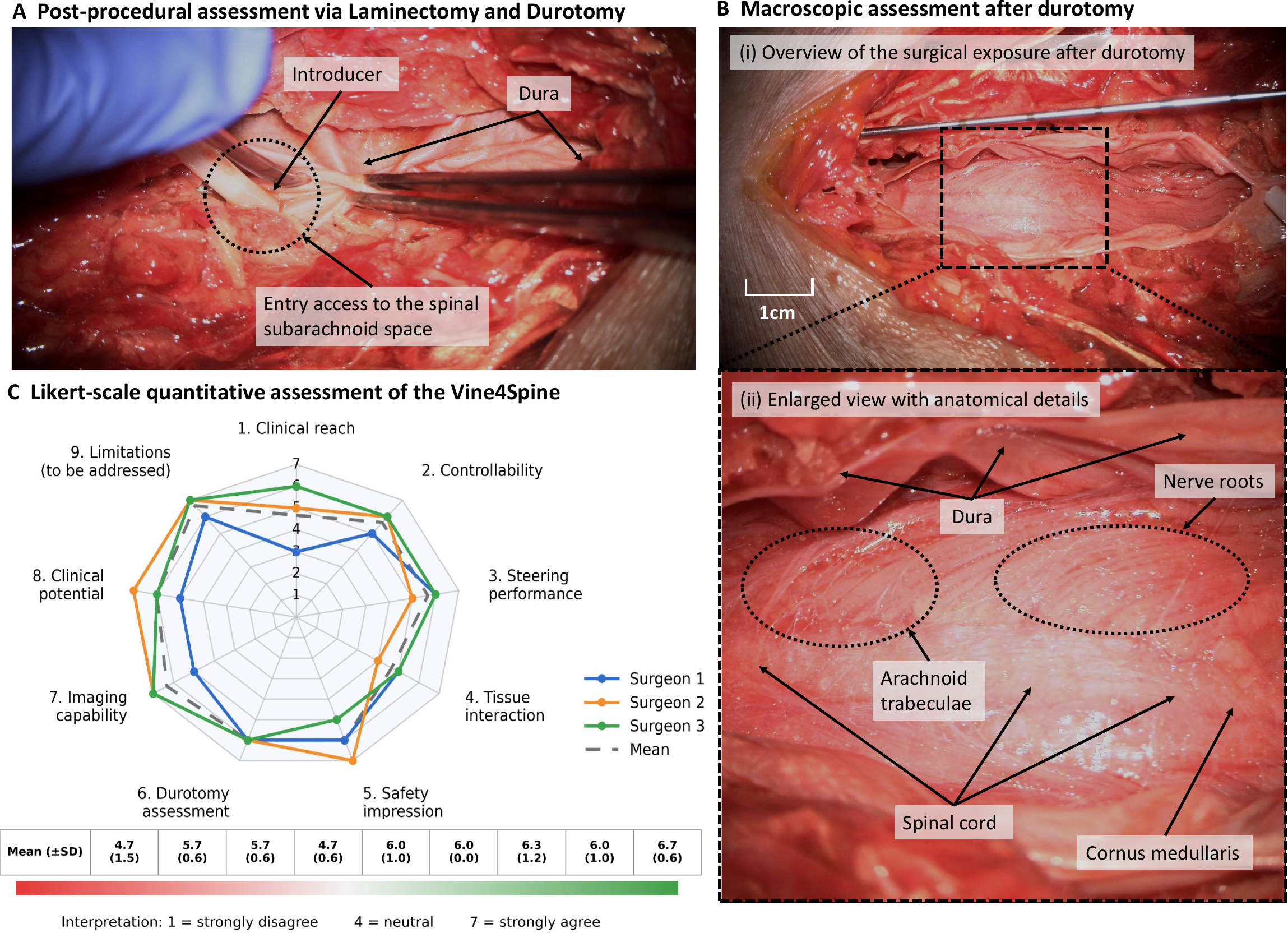}
    \caption {\textbf{Post-procedural neurosurgical assessment and quantitative clinical evaluation following the intrathecal navigation with Vine4Spine.} 
     \textbf{(A)} Macroscopic assessment after laminectomy and durotomy identifying the introducer and the entry site into the spinal subarachnoid space after removal of the posterior vertebral elements.
     \textbf{(B)} Intraoperative views of the exposed spinal subarachnoid space after durotomy.
     \textbf{(i)} Overview of the surgical field following a $5\,$cm durotomy.
     \textbf{(ii)} Enlarged view of the boxed region in \textbf{B(i)} showing preserved anatomical structures, including the dura mater, arachnoid trabeculae, spinal cord, nerve roots and conus medullaris. No visible macroscopic tissue injury was observed.
     \textbf{(C)} Quantitative clinical evaluation of the Vine4Spine system by three consultant neurosurgeons using a 7-point Likert-scale questionnaire. Individual ratings and the mean score are presented.
     }\label{fig.post_assessment}
\end{figure}

To further evaluate the safety during robotic navigation, the histological assessment was conducted via Laminectomy and Durotomy by neurosurgeons [Fig.~\ref{fig.post_assessment}A] (\textit{Supplementary Video10}). Post-procedural exposure of the spinal canal enabled direct inspection of the dura and underlying neural structures [Fig.~\ref{fig.post_assessment}B(i)]. No macroscopic disruption of the dura mater was observed following robotic navigation. After durotomy, the exposed neural elements and surrounding tissues appeared visually intact, with no evident signs of tearing, deformation, or contusion. The subarachnoid structures maintained their gross anatomical appearance throughout the inspected region, as demonstrated in  Fig.~\ref{fig.post_assessment}B(ii). These observations provide qualitative evidence that the navigation process did not result in apparent macroscopic tissue injury under the conditions tested.

\section*{Discussion}\label{sec:discussion}
\subsection*{Load-decoupled growth via tip-eversion establishes a new paradigm of intrathecal navigation}
Conventional intrathecal navigation relies on proximal pushing, inevitably transmitting compressive loads and friction along the entire catheter shaft as insertion depth increases. The present study demonstrates an alternative navigation paradigm in which robot extension is generated exclusively at the distal tip through eversion, fundamentally changing how mechanical loads are transferred to the surrounding anatomy. Rather than reducing interaction through increased flexibility or compliant materials alone, Vine4Spine eliminates translational sliding of the deployed body, thereby decoupling distal advancement from proximal pushing.

This work therefore extends the concept of tip-growing robots beyond proof-of-concept demonstrations by establishing, for the first time, their feasibility within the intact human spinal subarachnoid space.
Controlled intrathecal extension for a total robot growth of $150\,$mm was achieved under fluoroscopic and in-situ endoscopic guidance, with preservation of sheath-instrument alignment, steering authority, and sealing integrity throughout repeated duty-cycle transitions. 
The achieved intrathecal extension of $150\,$mm reflects clinically relevant access within the spinal canal, spanning multiple vertebral levels from a standard lumbar entry point. This range is sufficient to reach key anatomical regions of interest and is comparable to insertion lengths used in current intrathecal catheterisation. The present extension length was determined by the current hardware configuration, including the available sheath length and system integration constraints. Nevertheless, the eversion-based mechanism is inherently scalable, and longer extension lengths can be achieved in future developments through modifications to the storage chamber and sheath design.

Navigation within the spinal subarachnoid space presents a uniquely demanding mechanical environment: a narrow, compliant, fluid-filled corridor densely populated by fragile neural elements. Conventional push-based catheters propagate axial loads from the proximal base to the distal tip, accumulating friction along the deployed shaft and degrading distal controllability as insertion length increases. In contrast, the eversion-based architecture localises growth to the distal boundary. Because the deployed sheath remains stationary relative to surrounding anatomy, shear forces along the shaft are minimised and axial load propagation is reduced. 

The cadaveric results demonstrate that this load-decoupled growth principle is not limited to benchtop or phantom environments but can be sustained in intact human anatomy under clinically realistic access conditions. The preservation of steering curvature transmission at extended intrathecal lengths indicates that steering does not degrade with deployment depth, addressing limitations of push-based systems in compliant intraluminal environments.

Together, these findings demonstrate that load-decoupled growth is not restricted to benchtop environments but can be translated to clinically relevant spinal anatomy under conventional lumbar access.

\subsection*{Friction-dominated mechanics govern miniature eversion-based navigation}

The optimisation of duty-cycle parameters revealed that eversion-based systems operating at a miniature scale are governed by friction-dominated contact mechanics rather than idealised kinematic symmetry. Under purely geometric assumptions, a retraction-to-advancement ratio of $0.5$ would be expected to restore the relative alignment between the inner instrument and the everting sheath after each duty cycle. However, experimental observations demonstrated persistent residual coupling even when the internal pressure approached atmospheric conditions, indicating that friction hysteresis and membrane interactions within the folded sheath substantially influence the deployment mechanics.

Reducing the retraction ratio to $0.45$ and introducing sub-atmospheric vacuum pressure during the depressurisation phase effectively compensated for these friction-induced asymmetries, enabling sufficient mechanical decoupling and stable incremental growth over repeated duty cycles. In this context, duty-cycle parameters function not as purely geometric quantities, but rather as control variables that compensate for configuration-dependent frictional interactions between the folded sheath and the inner instrument.

Negative pressure plays a central role in this mechanism. By generating a pressure differential across the sheath wall, the depressurisation phase increases contact between the folded sheath layers and suppresses unintended sheath motion during instrument retraction. This stabilises the position of the deployed sheath while allowing the inner instrument to retract independently, thereby preserving coaxial alignment throughout successive growth cycles. Rather than serving solely as a mechanism for resetting the growth cycle, pressure regulation actively controls the mechanical coupling between the sheath and the inner instrument.

The computational model further supports this interpretation by demonstrating the dynamic evolution of interface forces throughout each duty cycle. While interaction forces increased during pressurisation and forward advancement, they progressively decreased during vacuum-assisted retraction, consistent with the experimentally implemented four-phase control strategy. These results indicate that pressure regulation serves not only as a growth actuator but also as a mechanism for controlling tissue interaction throughout the insertion cycle.

Together, these findings establish that stable eversion-based navigation in confined neural anatomy is possible with controlled internal pressure modulation of everting robots.

\subsection*{Structural integrity preserves mechanical and optical stability}

An important finding from the cadaver study is that the load-decoupled eversion architecture maintained its structural integrity throughout repeated deployment and steering cycles. No fluoroscopic evidence of contrast leakage, chamber collapse, or interruption of eversion was observed during navigation, indicating that the pressurised chamber remained effectively sealed despite repeated pressure cycling within the confined intrathecal environment. Together with the preserved coaxial alignment between the inner instrument and the everting sheath demonstrated by both fluoroscopic and endoscopic imaging, these observations indicate that the proposed architecture maintains structural stability over clinically relevant deployment lengths.

Maintaining instrument–sheath alignment is particularly important for pressure-driven growing robots because misalignment between the steering instrument and the deployed sheath may compromise both distal steering authority and optical performance. The consistent correspondence between fluoroscopic observations and endoscopic visualisation throughout deployment suggests that the depressurisation phase of the duty cycle effectively decouples motion between the inner instrument and the deployed sheath, preventing unintended sheath displacement while preserving the structural configuration required for continued navigation.

Beyond its mechanical role, this structural stability directly benefits in situ imaging. The embedded chip-on-tip endoscope relies on a stable geometric relationship between the imaging sensor, the surrounding sheath, and the cerebrospinal fluid interface. Transient sheath recoil or mechanical perturbations during repeated duty-cycle operation could introduce camera tilt, lens–sheath contact, or motion-induced blur, reducing image quality and navigation confidence. By suppressing unintended sheath motion and maintaining coaxial alignment, the proposed pressure-driven deployment strategy provides a stable optical environment that preserves a consistent field of view throughout intrathecal navigation.

Importantly, these observations suggest that pressure regulation fulfils multiple functions beyond driving distal growth. In addition to enabling eversion, the duty-cycle strategy preserves the structural integrity of the robot, stabilises the imaging platform, and maintains alignment between the steerable instrument and the deployed sheath during prolonged navigation. The ability to simultaneously maintain mechanical robustness, pressure sealing, steering capability, and endoscopic image quality represents an important systems-level advantage of the load-decoupled eversion architecture for minimally invasive intrathecal interventions.




\subsection*{Mechanical considerations for successful entry into the spinal subarachnoid space}
A critical phase of intrathecal navigation occurs immediately after the robot emerges from the rigid introducer and enters the spinal subarachnoid space. Unlike navigation within the introducer, where the robot is mechanically constrained, the distal segment becomes unsupported upon exiting the introducer and must simultaneously establish its orientation while maintaining sufficient structural stability for continued growth. Consequently, the first few millimetres of free distal extension represent the most mechanically demanding stage of deployment.

Successful entry requires a careful balance between tendon-driven steering and pressure-driven distal extension. If forward growth dominates before adequate steering is achieved, the distal segment experiences predominantly axial compressive loading, increasing the likelihood of tip buckling and deviation from the desired navigation axis. Conversely, excessive or delayed steering shifts the bending moment proximally, increasing mechanical loading near the introducer exit and reducing the effectiveness of distal curvature generation. The successful navigation demonstrated in this study suggests that appropriately coordinated steering immediately after introducer exit is essential for establishing the correct intrathecal trajectory while maintaining load-decoupled growth.

Once the robot is aligned with the axis of the spinal subarachnoid space, subsequent pressure-driven eversion substantially reduces axial loading along the deployed body, allowing distal curvature control to be preserved throughout continued navigation. This transition from a mechanically demanding entry phase to stable load-decoupled growth highlights the importance of coordinating steering and deployment during the initial stages of intrathecal access.

\subsection*{Biomechanical mechanisms underlying reduced tissue interaction}\label{sec:interaction mechanism}

Although the phantom experiments demonstrated substantial reductions in interaction forces during navigation, force measurements alone cannot explain how the eversion mechanism alters tissue loading. In particular, the comparison among continuous insertion, duty-cycle insertion and Vine4Spine showed that, while the duty-cycle strategy reduced the overall interaction force, it retained transient force peaks comparable to conventional insertion. In contrast, Vine4Spine simultaneously reduced both the mean and peak interaction forces and produced the smoothest force profile throughout navigation. These observations suggest that the reduced interaction cannot be explained solely by differences in insertion kinematics, but instead reflects a fundamentally different mode of mechanical interaction with the surrounding tissue.

Finite element analysis provides complementary biomechanical insight by revealing quantities that cannot be directly measured experimentally, including distributed stress fields, interfacial shear stress and contact area. The computational results indicate that the reduction in tissue loading arises primarily from redistribution of contact rather than simply lower insertion force. Compared with conventional catheter insertion, Vine4Spine generated a substantially larger contact interface while reducing both peak von Mises stress and interfacial shear stress throughout the surrounding tissue. Consequently, mechanical loads were distributed over the everted sheath instead of being concentrated at the advancing catheter tip. This load-decoupling behaviour provides a plausible mechanical explanation for the experimentally observed reductions in both peak and average interaction forces, as well as the suppression of the sustained loading and repeated force excursions characteristic of push-based navigation. 
It should be noted, however, that the contact mechanics quantified in the present study were obtained under dry phantom conditions. In the physiological intrathecal environment, cerebrospinal fluid is expected to provide additional lubrication at the robot–tissue interface, reducing frictional shear forces and associated shear stresses relative to those reported here. Consequently, the present measurements and simulations are likely to represent a conservative estimate of tissue loading during in vivo navigation, while the underlying load-redistribution mechanism of eversion-based growth remains unchanged.

Importantly, the finite element model reproduced the experimentally observed reduction in interaction force with good agreement for Vine4Spine, supporting its use as a tool for investigating tissue-level biomechanical phenomena beyond the capability of load-cell measurements alone. Although discrepancies remained in the predicted peak force during conventional catheter insertion, the model consistently captured the relative biomechanical advantages of eversion-based navigation. Together, the computational and experimental findings indicate that the principal benefit of Vine4Spine extends beyond reducing insertion force alone. By localising motion to the growing tip while the deployed sheath remains stationary relative to the surrounding anatomy, the system redistributes tissue loading, limits frictional shear accumulation along the robot body, and may therefore reduce the risk of localised mechanical trauma during navigation.

\subsection*{Translation towards clinically relevant intrathecal procedures}
Initial cadaveric investigations highlighted that the access interface between the robotic system and the spinal canal is a critical determinant of translational feasibility. Early attempts to deploy the system directly through a standard introducer for lumbar puncture revealed substantial friction at the entry interface, exacerbated by the viscosity of contrast-containing pressurisation media and the tight dimensional tolerances of the introducer lumen. These factors increased resistance to sheath eversion and limited the reliability of initial deployment. 

To address this limitation, a clinical introducer (Cordis AVANTI®) was adopted as the access interface. The introducer provided a large, smooth internal lumen and a sealed side port, enabling controlled saline flushing throughout deployment. Continuous or intermittent saline irrigation maintained a fluid-filled pathway analogous to the CSF environment, reducing friction at the access corridor and facilitating eversion without risk of rupture. This configuration preserved compatibility with standard lumbar puncture workflow and required no additional surgical exposure.

Navigation experiments further revealed that precise regulation of the intrathecal extension length upon emergence from the introducer is critical. At the L$4$–L$5$ level, the dural sac provides a compliant but spatially constrained corridor, where neural elements such as the cauda equina occupy a significant proportion of the available volume. 
Consequently, even small deviations in deployed length or orientation may result in off-axis contact with neural or vascular structures. To address this, the initial $2-3\,$mm of the robot needs high steering capability to enable generation of a curvature that facilitates tangential alignment with the spinal canal for subsequent sheath growth. 

Together, our findings illustrate that intrathecal navigation using tip-growing robots requires coordinated regulation of extension length, steering actuation, and access interface conditions. While the eversion-based growth mechanism mitigates friction accumulation along the deployed shaft, effective clinical deployment still depends on maintaining a low-resistance entry interface and carefully managing the mechanical transition from introducer-guided insertion to free intrathecal navigation.

\subsection*{Safety implications across computational, phantom and cadaver validation}
The present study evaluates navigation safety using three complementary levels of evidence. Computational modelling demonstrated substantial reductions in local tissue stress and interfacial shear stress; phantom experiments quantified significantly lower global interaction forces during navigation; and cadaveric studies revealed no observable macroscopic disruption of the dura mater or surrounding neural structures following navigation. Together, these independent observations consistently support the hypothesis that eversion-based navigation reduces tissue loading compared with conventional push-based insertion.

Nevertheless, the present safety assessment remains preliminary. Computational predictions were obtained using a phantom-based material model, phantom experiments quantified global rather than local tissue forces, and cadaveric evaluation relied on qualitative post-procedural inspection without histological analysis. Consequently, although the combined evidence supports the biomechanical advantages of load-decoupled navigation, further studies incorporating histopathological assessment, quantitative cadaveric force measurement and physiological cerebrospinal fluid conditions will be required before definitive conclusions regarding clinical safety can be drawn.

The absence of observable macroscopic tissue damage following post-procedural exposure suggests that the eversion-based navigation mechanism operates with a low level of mechanical interaction in the confined spinal subarachnoid space. This observation is consistent with the reduced interaction forces quantified in phantom experiments, supporting the premise that tip-based growth may mitigate tissue loading compared to conventional push-based insertion strategies. This enabled the clinical experts to comment not only on usability and workflow, but also on visible tissue interaction, local trauma, dural disruption, arachnoid disturbance, and nerve-root displacement compared with conventional push-based access.

Macroscopic post-procedural inspection by neurosurgeons represents an appropriate endpoint at this prototype validation stage, in accordance with preclinical device evaluation frameworks. However, it is qualitative and limited to visual inspection by neurosurgeons, with direct force measurement in cadaveric conditions and histological analysis as key next steps in the safety evaluation of the navigation approach. 

\subsection*{Clinical Perspectives on Vine4Spine}
To obtain an initial end-user evaluation of the proposed platform, three consultant neurosurgeons who participated in the cadaveric studies completed a structured 7-point Likert-scale assessment comparing Vine4Spine with conventional catheter-based approaches across multiple domains, including controllability, tissue interaction, imaging capability, navigation performance, and overall usability (Fig.~\ref{fig.post_assessment}C). 
In addition, participants completed questionnaires and provided qualitative feedback on the system's perceived strengths, limitations, and potential clinical applications. Two of the three neurosurgeons also participated in the post-procedural anatomical inspection.

Across the evaluated domains, Vine4Spine received consistently favourable ratings, suggesting a positive expert perception of its potential utility for minimally invasive spinal interventions. Notably, the highest-rated attributes were associated with navigation and visualisation capabilities, reflecting the system's ability to combine steerable intrathecal access with direct endoscopic imaging. These functionalities address recognised limitations of conventional catheter-based techniques, which typically rely on indirect imaging and provide limited visual feedback during navigation.

Qualitative feedback further reinforced these observations. Participants consistently highlighted the combination of minimally invasive access, navigational capability within the spinal subarachnoid space, and direct visualisation of anatomical structures as the most distinctive features of the platform. One neurosurgeon remarked that “the catheter-based ones are blind, whereas this system allows you to navigate up the spine and visualise structures”, underscoring the perceived value of integrating navigation and in situ imaging within a single robotic system.

Although the present evaluation involved a limited number of participants, all evaluators were consultant neurosurgeons with direct experience in spinal procedures and represented the intended end-user group. Consequently, the quantitative assessments and qualitative feedback should be interpreted as an expert clinical appraisal of the technology rather than a population-level assessment of clinical acceptance. Future studies involving larger clinician cohorts across multiple centres will be necessary to further evaluate usability, workflow integration, and clinical utility.

The positive expert feedback should be interpreted alongside the biomechanical findings presented in this study. Rather than evaluating only usability, clinicians consistently recognised the combination of steerable navigation, direct visualisation and reduced tissue interaction as distinguishing characteristics compared with conventional catheter-based techniques. Although limited in scale, these observations provide preliminary clinical support for the proposed navigation paradigm and motivate future multi-centre evaluation involving a larger cohort of end users.

\subsection*{Limitations and outlook}

Our preliminary investigation across two cadaveric specimens enabled optimisation of the growth and steering parameters, as well as the overall imaging and robotic architecture. However, the principal findings of this study were demonstrated in a single fresh-frozen cadaver. Inter-subject variability, including differences in spinal canal geometry, CSF distribution, and pathological conditions, may influence navigation performance and deployment behaviour, and therefore warrants further validation across a larger cohort of specimens. In addition, the cadaveric specimen did not provide physiological pressure dynamics. While on one hand, the freezing/thawing and overall fluid degradation made visualisation more challenging, dynamic fluidic effects on growth and steering were not captured.

Quantitative measurement of interaction forces was not performed during cadaveric deployment. Consequently, safe access was inferred from the reduced interaction forces observed in phantom benchmarking, in conjunction with macroscopic observation under real-time fluoroscopic and endoscopic imaging, and post-procedural microscopic inspection following Laminectomy and Durotomy. While no abnormalities were observed following post-procedural exposure, the present assessment of tissue integrity remains limited to qualitative visual inspection. The absence of direct evidence of tissue disruption at the histological level precludes definitive conclusions regarding subtle or subclinical injury. In particular, visual inspection following Laminectomy and Durotomy is inherently insensitive to microstructural alterations, including localised cellular damage, inflammatory responses, or minor vascular compromise. 

Several limitations are specific to the computational modelling framework. The finite element model was developed using the experimentally validated phantom geometry rather than the donor-specific spinal anatomy and therefore should be interpreted as a mechanistic model for investigating tissue loading rather than a predictive simulation of in vivo behaviour. While the model reproduced the experimentally observed reduction in global interaction force for Vine4Spine, direct validation of the predicted stress redistribution, contact area and interfacial shear stress remains beyond the capability of the current experimental methodology. Consequently, these local biomechanical quantities should be interpreted as computationally supported mechanisms rather than directly measured experimental observations.
Furthermore, constitutive properties and friction coefficients were derived from experimental characterisation of the phantom rather than native spinal tissues. Uncertainties in these parameters, together with the simplified representation of the intrathecal environment, are likely to contribute to discrepancies between simulated and measured forces, particularly for conventional catheter insertion. Future work will incorporate patient-specific anatomical reconstruction, experimentally measured tissue constitutive properties and tribology, cerebrospinal fluid–structure interaction, and multimodal experimental validation of local tissue deformation and stress to improve the physiological fidelity and predictive capability of the computational framework.

Finally, the present system is a research prototype. Further engineering development is required to optimise device performance, improve the robustness of the control, ensure a longer eversion length, a therapy delivery channel, and capabilities of real-time estimation of interaction force. Translation towards clinical application will require compliance with relevant medical device standards and validation in physiologically representative environments.

Despite these limitations, the present work establishes eversion-based robotic growth as a viable strategy for extended navigation within the spinal subarachnoid space. By decoupling distal advancement from proximal pushing, the system preserves steering authority, reduces distributed frictional interaction, and maintains mechanical–optical stability within a confined and delicate neural environment.

\section*{Materials and Methods}
\subsection*{System Development and Overall Architecture}
\subsubsection*{Eversion Growing System}
The eversion growing system is composed of a pressurisation chamber, an active channel, an eversion growing sheath, and an inner-placed instrument that offers steering, interaction and visualisation functionality. 
The pressurisation chamber accommodates the pressurisation medium and the eversion growing sheath. Its inner pressure is regulated to control robot growth.

In our benchtop tests, we used saline due to its low compressibility and biocompatibility. From a safety perspective, saline leakage into the spinal canal is medically acceptable. 
During our cadaver study, a diluted contrast agent was employed as the pressurisation medium to help visualise the eversion growing sheath independently of the inner placed metallic instrument under fluoroscopy. Specifically, this approach helped identify the sheath tip location and quantify whether leakage or misalignment occurred during navigation. The contrast agent employed was Omnipaque\textsuperscript{TM} X-ray contrast medium (GE HealthCare, USA), with an iodine concentration of $240$\,mg I/mL and diluted in saline at a contrast-agent-to-saline ratio of $3:4$. This dilution ratio maintained the efficient visualisation of the sheath while maintaining an appropriately low viscosity of the pressurisation medium. The visibility of the mixture using the identified dilution ratio was evaluated by pre-scanning a range of concentrations under fluoroscopy during preliminary experiments. 

The growing sheath was made of sealing-less PTFE liners (Zeus Industrial Products Inc., USA), with an outer diameter (OD) of $2.0\,$mm and $20\,\mu$m wall thickness. PTFE is a biocompatible and clinically approved material. 
The active channel, to which the proximal tip of the eversion growing sheath connects, was a stainless steel tube with an OD of $2.0\,$mm and an inner diameter (ID) of $1.8\,$mm, motorised to implement the release and pull of the sheath stored in the chamber.

The distal tip of the eversion growing sheath was reversed and connected with the tip nozzle of the eversion growing chamber. The tip nozzle of the pressurisation chamber was customised using a $14$ Gauge syringe needle. Luer locks were used to enable rapid assembly and disassembly. To reduce the friction forces exerted during eversion of the sheath, its original tip was replaced by a $5\,$mm-long NiTi tube (ID: $1.9\,$mm; OD: $2.0\,$mm), with an ultra-thin thickness of $50\,$$\mu$m.

The pressurisation chamber that accommodated the entire eversion growing element was made of transparent acrylic tubing (ID: $3.0\,$mm; OD: $6.0\,$mm) (RS Components Ltd, UK). Two M$4$ push-fit components (Festo Ltd., UK) were integrated onto drilled ports on the chamber, connecting to M$4$ silicone tubing for both real-time pressure measurement and the pressurisation medium management to achieve pressure regulation of the chamber. A GateWay PLUS Y-adapter (Boston Scientific Corporation, USA) placed at the rear of the pressurisation chamber ensured its sealing despite axial translation of the active channel.

\subsubsection*{Inner-placed instrument for steering and imaging}

The presence of the inner-placed instrument necessitates duty-cycle control of the eversion process at the benefit of accommodating functional components for imaging, steering, and, in the future, therapy delivery injectors. It was composed of a NiTi steerable segment, a miniature camera module, an illumination fibre, a NiTi tendon, the instrument shaft and the covering sheath, as specified in the Supplementary Materials.

Via precise laser patterning (PeierTech, China), notches were fabricated onto the front NiTi segment to form a $100\,$mm-long steerable segment (ID: $1.0\,$mm; OD: $1.2\,$mm). A NiTi tendon passing through its central lumen was attached to its tip, and enabled steering up to a maximum bending curvature of up to $6.28\mathrm{mm^{-1}}$. 

A NanEyeM miniature chip-on-tip camera module (ams-OSRAM AG, Austria) was fixed with UV-Curable optical adhesive onto the tip of the steerable segment, which is transparent for illumination spreading. The chip's $1\,\mathrm{mm^2}$-surface footprint offers a $320\times320$ resolution, based on sensitive $2.4\,$ $\mu$m rolling-shutter pixels with large full-well capacitance, optimised for medical endoscopic imaging where high Signal-to-Noise Ratio (SNR) is essential. A flat 3-meter-long imaging cable, with dimensions $1.2\,$mm $\times$ $0.2\,$mm was accommodated exterior to the inner-placed instrument, opposite to the notched side, ultimately connecting to a NanEye Fibre Optic Box (ams-OSRAM AG, Austria). An ultra-thin PTFE sheath (ID: $1.42\,$mm; OD: $1.46\,$mm) (Zeus, USA) covered the imaging cable to protect it and maintain reduced surface friction when the inner instrument was moving within the eversion growing sheath.

Illumination was provided by an optical fibre traversing the robot's full central lumen up to $5\,$mm posterior to the camera tip, where it was adhered with Norland Optical Adhesive (Edmund Optics Inc., USA), which is a transparent UV-curable chemical. The fibre had a coated diameter of $0.50\,$mm, a length of $\sim 160\,$cm with sufficient redundancy, and was connected to a Fibre-Coupled LED (Thorlabs, USA) via a bare fibre optic terminator (Thorlabs, USA) powered by a T-Cube\textsuperscript{TM} LED Driver (Thorlabs, USA). The illumination optical fibre offers illumination with a power up to $27\,$mW, substantially improving light penetration within the fluid-filled intrathecal space.
Such a configuration was pre-tested in our phantom study to ensure that illumination and image acquisition do not interfere with robot navigation. 
The overall architecture is illustrated in the Supplementary Materials.

\subsection*{Duty cycle implementation with pressure regulation}

To achieve stable, controllable tip-based growth while maintaining alignment between the eversion sheath and the inner-placed steerable instrument, Vine4Spine employed a four-stage duty-cycle control strategy, which decoupled pressure-driven eversion from internal instrument translation, thereby enabling precise growth within confined, compliant anatomy.


\textit{Phase I: Sheath Pressurisation (t$1$):}
At the beginning of the duty cycle, the eversion growing sheath was pressurised to a high pressure level ($P_h$) within the pressurisation chamber. During this stage, the active channel and inner instrument remained stationary. The applied pressure generated axial tension in the folded sheath and prepared the system for eversion by overcoming static friction at the distal nozzle. No net extension occurred at this stage.

\textit{Phase II: Forward Advancement (t$2$):}
Once the target pressure $P_h$ was reached, coordinated advancement of the system was initiated. The inner-placed instrument advanced forward, while the eversion sheath everted at the distal nozzle and grew at approximately half the translational velocity of the catheter. This differential motion enabled controlled deployment of new sheath material at the tip while preserving alignment between the sheath tip and the inner instrument. As a result, the robot length increased without inducing relative sliding along the already deployed sheath.

\textit{Phase III: Depressurisation (t$3$):} After the desired growth increment was achieved, the internal pressure was reduced to a lower pressure level ($P_l$). This depressurisation halted further eversion and stabilised the newly deployed sheath segment. During this phase, no forward growth occurred, and the system transitioned into a mechanically stable state in preparation for realignment.

\textit{Phase IV: Catheter Retraction (t$4$):}
In the final stage, the inner-placed catheter retracted proximally while the eversion sheath remained stationary. This motion realigned the instrument tip with the sheath tip, restoring the initial relative configuration prior to the next duty cycle commencing. Accurate realignment was critical to ensure consistent steering authority, imaging alignment, and repeatable growth in subsequent cycles.

By repeating this four-stage sequence, Vine4Spine achieved incremental, pressure-controlled tip growth while avoiding continuous axial loading, sheath buckling, or uncontrolled frictional interaction with surrounding tissue. The duty-cycle strategy allowed growth, steering, and sensing to be coordinated in a predictable manner, enabling safe navigation within the spinal subarachnoid space where conventional push-based insertion is mechanically constrained. Moreover, the modelling details of its workspace is presented in the Supplementary Materials.

\subsection*{Duty-cycle parameter definition and optimisation}
The duty-cycle parameters were defined and optimised through free-space deployment experiments before phantom and cadaveric evaluation. During the forward-stroke phase, the inner instrument was advanced by a commanded displacement $\Delta x_{adv}$, while coordinated translation of the active channel enabled the sheath to evert and extend at approximately half the translational rate of the instrument. During the subsequent retraction phase, the inner instrument was displaced proximally by $\Delta x_{ret}$. The retraction-to-advancement ratio was defined as:
\begin{equation*}
    r = \frac{\Delta x_{ret}}{\Delta x_{adv}}
\end{equation*}
Under ideal kinematic conditions, a ratio of $r=0.50$ would be expected to restore alignment between the instrument and sheath tips because the sheath extends by approximately half the instrument advancement during eversion. However, preliminary experiments showed that residual frictional coupling and hysteresis between the instrument and folded sheath produced non-ideal relative motion during depressurisation and retraction.

To identify parameters that maintained instrument–sheath alignment over repeated cycles, low-level chamber pressure $P_l$ was varied from $80$ to $110\,$kPa in increments of $10\,$kPa, in combination with retraction ratios of $r=0.40$ and $r=0.45$. For each parameter combination, the robot was deployed in free space over its full available extension of $150\,$mm. Final instrument–sheath misalignment was quantified as the axial distance between the distal end of the inner instrument and the distal end of the everted sheath after full deployment. A nominal ratio of $r=0.50$ was also evaluated during preliminary tests but was excluded from the final parameter comparison because repeated operation produced progressive instrument lag, distal sheath accumulation and eventual sheath rupture.

During the low-pressure phase, chamber pressure was reduced to the prescribed sub-atmospheric level $P_l$ to reduce instrument–sheath coupling before instrument retraction. The high-level pressure was maintained at $P_h =140\,$kPa throughout the optimisation experiments. The combination $P_l=90\,$kPa and $r=0.45$ produced the smallest final instrument–sheath misalignment and enabled stable deployment over the full $150\,$mm extension. These parameters were therefore used in all subsequent phantom and cadaveric experiments.

\subsection*{Finite element modelling of robot–tissue interaction}
To reveal the biomechanical mechanisms underlying the experimentally observed interaction reduction achieved by Vine4Spine, a nonlinear finite element model was developed using the explicit dynamics solver LS-DYNA\textregistered (Ansys, Inc.; MPP Version R14.1.1). Unlike the phantom experiments, which provided only global interaction forces through an external load cell, the computational model enabled the quantification of spatially distributed biomechanical quantities, including contact area, von Mises stress, interfacial shear stress, and interface forces throughout insertion. Two insertion strategies were simulated under identical conditions: conventional push-based catheter insertion and the Vine4Spine duty-cycle insertion protocol. Model fidelity was assessed by comparing the simulated insertion forces with those measured experimentally.

The spine phantom was represented using a visco-hyperelastic constitutive model to reproduce the finite deformation behaviour of the Agilus30-based phantom material observed experimentally. The PTFE eversion sheath was modelled as a linear elastic material, while the catheter and steering tendon were also represented using linear elastic constitutive models. Although these components were fabricated from Nitinol, kinematic analysis confirmed that the maximum strains experienced during the prescribed duty-cycle motion remained below the onset of stress-induced martensitic transformation, allowing a linear elastic approximation without affecting the predicted interaction mechanics. Material parameters are detailed in the Supplementary materials.

Mechanical interactions between the robotic system and surrounding tissues were described using a symmetric penalty-based surface-to-surface contact formulation with literature-informed Coulomb friction coefficients for the catheter–phantom, sheath–phantom, and catheter–sheath interfaces. A soft-constraint penalty formulation was employed to ensure stable contact enforcement between the compliant phantom and comparatively stiff robotic components while avoiding excessive numerical penetration. Complete contact definitions and friction parameters are provided in the Supplementary Information.

Boundary conditions replicated the experimental insertion protocols for both navigation strategies. Conventional catheter insertion was implemented as continuous forward advancement over the same insertion distance as in the experiments. Vine4Spine was modelled using the four-stage duty-cycle strategy consisting of sheath pressurisation, coordinated catheter and sheath advancement, vacuum-assisted sheath release, and proximal catheter realignment. Pneumatic loading and translational boundary conditions were synchronised throughout each cycle to reproduce the experimental pressure-controlled eversion process. To improve numerical stability, all prescribed displacement trajectories were generated using smooth cosine-interpolated motion profiles, avoiding abrupt velocity reversals and the artificial inertial loading associated with discontinuous boundary conditions. The resulting kinematics reproduced the experimentally implemented duty-cycle insertion while maintaining stable explicit time integration.

Simulated insertion forces were compared with phantom experiments for model validation, whereas the remaining biomechanical quantities were used to investigate the mechanical mechanisms responsible for the reduced tissue interaction achieved by Vine4Spine. Detailed descriptions of the computational geometry, constitutive models, contact definitions, mesh discretisation, friction parameters, and boundary conditions are provided in the Supplementary Information.

\subsection*{Benchmarking of Vine4Spine with a spine phantom}
\subsubsection*{Spine phantom for navigation evaluation}
A bespoke cervical spine phantom was developed to provide an anatomically representative and optically accessible platform for evaluating eversion-based robotic navigation prior to cadaveric studies.
The phantom is made of the Agilus30\texttrademark Polyjet Photopolymer (Stratasys, USA) in clear to represent the dura mater, central canal, and paired nerve bundles,
and was reconstructed from open-source T$2$ MRI anatomical imaging data\cite{chierichini2016cervical}. Individual anatomical components were cast in soft platinum-cure silicone and assembled within a transparent outer dura, allowing physiologically relevant deformation while maintaining direct visual access during robot deployment. Three predefined navigation targets were incorporated within the spinal canal to provide reproducible steering objectives throughout the experiments. A custom introducer was mounted at a clinically representative insertion angle to emulate lumbar access into the spinal subarachnoid space.

\subsubsection*{Phantom navigation protocol} 
The Vine4Spine system was introduced through the custom introducer and deployed using the optimised duty-cycle parameters identified during the free-space optimisation experiments. Following emergence from the introducer, the steerable distal segment was aligned with the central axis of the spinal canal before incremental pressure-driven tip growth towards the predefined navigation targets. Navigation consisted of alternating steering and eversion-growth commands until the maximum available deployment length of $150\,$mm was reached. External video recordings and embedded endoscopic images were acquired simultaneously throughout navigation to assess deployment stability, steering performance, target acquisition and instrument–sheath alignment.

\subsubsection*{Endoscopic imaging evaluation} 
To evaluate the integrated imaging system under conditions representative of the intrathecal environment, the transparent spine phantom was enclosed within a pigmented sealed tube to eliminate ambient illumination. Saline mixed with milk and red food-colouring gel was used to reproduce the absorption, scattering and colour characteristics observed during preliminary cadaveric studies. Illumination was provided by the integrated optical fibre, while images were acquired using the embedded NanEye chip-on-tip camera. This experimental configuration enabled assessment of illumination performance, image quality and anatomical visualisation during simultaneous robot growth and steering.

\subsubsection*{Comparative navigation protocols} 
Three navigation strategies were evaluated under matched experimental conditions. 
\textbf{Conventional catheter insertion} was implemented as continuous proximal advancement of the steerable catheter. To enable a fair comparison with the duty-cycle-based approaches, the insertion velocity was adjusted such that the average translational velocity matched that of the Vine4Spine duty-cycle protocol. 
\textbf{Duty-cycle catheter insertion} was implemented by applying the same duty-cycle timing, cumulative insertion distance, and average translational velocity as Vine4Spine, but without pressure-driven sheath eversion. 
\textbf{Vine4Spine} navigation employed the proposed four-stage pressure-regulated duty-cycle strategy with coordinated distal sheath eversion and inner-instrument translation. Each navigation strategy was repeated three times using identical phantom geometry, insertion trajectory, insertion distance, and experimental boundary conditions.

\subsubsection*{Interaction force measurement in phantom experiments} 
Each of the above-mentioned navigation strategies was conducted three times, and a full plot of the force profile is available in the Supplementary Materials. Interaction forces were measured using a six-axis force/torque sensor (Mini40, ATI Industrial Automation, USA). The sensor was calibrated with the SI-40-2 calibration configuration. It provided a force measurement range of $±20\,$N with a resolution of $1/200\,$N along the X/Y axes, and a measurement range of $±60\,$N with a resolution of $1/100\,$N along the Z axis, separately. Force signals were sampled at $30$Hz through an ATI EtherCAT interface and recorded using Python for offline analysis. The sensor was mounted on the base of the spine phantom, with its X+ axis aligned with the axial direction of the spine to capture interaction forces during robot advancement. Prior to each trial, the sensor output was zeroed to remove static bias.

\subsubsection*{Force-data processing and quantitative analysis} 
Force data were analysed offline using MATLAB (MathWorks, USA). The onset of robot–phantom interaction was identified from the axial force ($F_x$), after which all recordings were temporally aligned by defining the contact point as t=0. Signals preceding the identified contact were discarded for comparative analysis. To remove measurement bias associated with sensor mounting, the baseline of the $F_z$ component was corrected by subtracting the mean of the first ten measurements following the beginning of the analysed interval. No baseline correction was applied to $F_x$ or $F_y$.

The measured force components were transformed into biomechanically meaningful quantities for comparison. The axial interaction force was defined as 
\begin{equation*}
 F_{axial}=F_x   
\end{equation*}
the transverse interaction force as
\begin{equation*}
 F_{transverse}= \sqrt{F_x^2+F_y^2}   
\end{equation*}
and the resultant interaction force $\|\mathbf{F}\|$ as
\begin{equation*}
 \|\mathbf{F}\|= \sqrt{F_x^2+F_y^2+F_z^2}   
\end{equation*}

For each trial, the mean absolute value and maximum value of the axial, transverse and resultant interaction forces were calculated. Results are reported as mean ± standard deviation across the three repeated experiments. Percentage reductions achieved by Vine4Spine relative to conventional insertion and duty-cycle insertion were calculated from the corresponding group means. For visual comparison, the radar plot was generated by normalising each force metric to the corresponding mean value obtained from conventional catheter insertion, which was assigned a value of $100\%$.

\subsection*{Surgical procedure and data processing}

\subsubsection*{Cadaver preparation and ethical approval}
Cadaver experiments were conducted under approval from the UK Human Tissue Authority (HTA; licence number $12778$), granted under Section $16$ ($2$) (d) of the Human Tissue Act $2004$. A fresh-frozen human cadaver (male, $92$ years old) with an intact central nervous system, including the entire spine, was thawed to room temperature ($\sim\SI{24}{\degreeCelsius}$) in a mock operating environment prior to experimentation.

Throughout the procedure, fluoroscopic imaging was performed using a Siemens Cios Spin C-arm system (Siemens, Germany) to guide intrathecal access, robot deployment and navigation. To enable independent visualisation of the everting sheath during fluoroscopy, the pressurisation medium consisted of saline mixed with a diluted Omnipaque\texttrademark (GE HealthCare, USA; iodine concentration $240\,$mg I/mL)contrast agent, with a contrast-to-saline volume ratio of $3:4$, as described above. 

\subsubsection*{Cadaveric surgical access and robotic deployment}

Neurosurgeons established intrathecal access following a standard lumbar puncture procedure. A Tuohy needle (Medtronic, USA) surrounded by a Cordis AVANTI\textregistered introducer sheath (Cordis, USA) was inserted percutaneously to access the spinal subarachnoid space under fluoroscopic guidance. Correct intrathecal positioning was confirmed by cerebrospinal fluid return together with fluoroscopic imaging. Following confirmation, the Tuohy needle was removed while the introducer remained in position to provide the access channel for robotic deployment.

Prior to deployment, sterile saline was flushed through the introducer to remove residual debris, maintain a lubricated fluid pathway, and minimise insertion resistance. Approximately $50\,$mm of the everting sheath was preloaded within the introducer to establish the initial deployment configuration. The Vine4Spine system was positioned adjacent to the operating table and aligned with the introducer to permit unobstructed deployment through the established access route.
During the cadaver experiments, duty-cycle navigation protocol was employed, as detailed above. Robot translation, steering and eversion were teleoperated using a joystick-based interface while fluoroscopic and endoscopic images were displayed simultaneously for real-time guidance.

Navigation consisted of incremental pressure-driven tip growth interleaved with tendon-driven steering adjustments until the maximum available deployment length of 150 mm was reached. Advancement was terminated when the predefined deployment length or anatomical target was reached, as determined jointly by the operating neurosurgeons and the robot operator.

To minimise X-ray exposure, all personnel wore lead protective suits and personal dosimeters, and a mobile lead shielding board was positioned between the robot operator and the fluoroscopy platform during imaging.

\subsubsection*{Fluoroscopic and endoscopic image acquisition}
Fluoroscopic images were acquired throughout deployment at $\sim30$ frames per second and exported from the Siemens Cios Spin system in DICOM format. Endoscopic imaging was simultaneously acquired using the embedded NanEye chip-on-tip camera. Raw endoscopic recordings were streamed as 16-bit Bayer-pattern (BGGR) image sequences using the NanEye Viewer software (v9.3.3.1, ams-OSRAM AG, Austria). Fluoroscopic and endoscopic recordings were synchronised manually based on deployment events for qualitative comparison throughout robot navigation.

\subsubsection*{Motion, steering and deployment analysis}
Robot deployment was analysed from fluoroscopic recordings using semi-automatic tracking of the distal tip within a manually selected region of interest. The reconstructed deployment trajectory was segmented into three growth segments separated by two steering adjustments. A filtered reference trajectory was generated from the tracked tip positions, and the point-wise deviation between the raw and filtered trajectories was calculated throughout deployment. This deviation was used to quantify the oscillatory motion introduced by periodic duty-cycle actuation and robot–environment interaction. Stage-wise distributions of the deviation were summarised using boxplots.

Following full deployment, steering performance was quantified from fluoroscopic recordings. Instrument centrelines corresponding to the initial, maximum and minimum steering configurations were reconstructed from manually identified centreline points, and image-plane curvature was calculated from the fitted centrelines. Distal tip motion during steering was quantified by semi-automatic optical-flow tracking combined with Kalman filtering. Tip displacement was measured relative to the initial steering configuration and converted from pixels to millimetres using fluoroscopic image calibration.

Fluoroscopic and endoscopic recordings were analysed concurrently to assess qualitative correspondence between robot motion and anatomical visualisation. Steering manoeuvres identified fluoroscopically were compared with the corresponding changes observed in the endoscopic field of view to confirm consistent distal orientation throughout deployment.

\subsubsection*{Instrument–sheath alignment and sealing assessment}
Instrument–sheath alignment was evaluated following full robot deployment by gradually retracting the inner instrument while the everting sheath remained stationary. Fluoroscopy was used to visualise the contrast-filled sheath as it became progressively exposed, while the embedded endoscope simultaneously visualised the internal sheath boundary. Alignment was assessed qualitatively from the correspondence between fluoroscopic sheath exposure and the endoscopic sheath contour, together with the absence of visible sheath displacement, collapse or distortion during instrument retraction.

Sealing integrity was evaluated by continuous monitoring of chamber pressure together with fluoroscopic inspection for visible contrast leakage. Before cadaveric deployment, the assembled robotic system was pressure-tested up to $190\,$kPa to verify sealing integrity above the nominal operating pressure. During the cadaver study, transient pressurisation up to $180\,$kPa was performed under closed-loop regulation. Structural integrity was considered preserved when no uncommanded pressure loss or fluoroscopic evidence of contrast dispersion outside the sheath was observed.



\subsubsection*{Post-procedural exposure and qualitative safety assessment}
Following completion of robot navigation, neurosurgeons performed a multilevel laminectomy to expose the dorsal dura mater. The dura was inspected for visible abnormalities before a longitudinal durotomy was carried out under microscopic guidance using a ZEISS KINEVO surgical microscope (ZEISS, Germany). Direct visual inspection of the spinal cord, arachnoid trabeculae, nerve roots and surrounding tissues was performed to identify macroscopic evidence of tearing, deformation, haemorrhage or other structural disruption. Observations were recorded immediately following exposure.



\subsubsection*{Data types and post-processing}
Fluoroscopic DICOM images were converted to PNG format using StarViewer (v0.14.0-RC1), and image sequences were reconstructed into videos using Python (v3.12.11) and the open-source \hyperref[https://ffmpeg.org]{FFmpeg} toolkit \cite{FFmpeg}.
Microscopic images and videos were acquired using a ZEISS KINEVO surgical microscope (ZEISS, Germany).
Raw endoscopic recordings were converted into 16-bit Bayer-format image sequences before post-processing using a Python-based pipeline integrating FFmpeg, OpenCV and ImageMagick. Colour balance, brightness, contrast, gamma correction, denoising, sharpening and temporal smoothing were interactively adjusted to optimise visualisation of intrathecal anatomical structures. Final videos were encoded using H.264 (CRF 10, yuv444p), and additional lossless FFV1 videos together with PNG frame sequences were exported for figure preparation and quantitative analysis.

\bibliography{sample}
\section*{Data Availability}
The authors declare that all data supporting the findings of this study are available within the paper and its supplementary information. 
The data generated in this study are provided in the main text, supplementary information, and source data file. Additional data are
available from the corresponding author on request. 
Source data are provided as a Source Data file in this paper.

\section*{Acknowledgements}
This work was supported by Innovate UK [SoftReach / 10062486], EPSRC [EndoTheranostics / EP/Z003172/1] under the Horizon Europe Guarantee Extension (ERC Synergy), and the European Union, under grant agreement No 101099145, project SoftReach. Work in the Surgical \& Interventional Engineering Validation Suite (mock OR) was supported by core funding from the Wellcome/EPSRC Centre for Medical Engineering [WT203148/Z/16/Z], a multi-user equipment grant for post-mortem evaluation of medical devices from Wellcome [218286/Z/19/Z], and the Wolfson Foundation [PR/ylr/md/21896]. The authors would like to acknowledge Mr Duane James (Specialist Chief Technician) and Miss Gayathri Nantharatnam (Technician) for their preparation and assistance with the cadaveric study. 

\section*{Author contributions}
ZW (Methodology, Software, Setup Development, Validation, Formal Analysis, Investigation, Data Curation, Writing), PK (Methodology, Simulation, Investigation), AA, JS, TB (Methodology, Investigation, Reviewing), SS (Methodology, Supervision), CBa (Methodology, Investigation), WF, SO (Reviewing, Funding Acquisition, Resources), PV (Methodology, Review), CB (Methodology, Investigation, Resources, Writing, Reviewing, Supervision, Project Administration, Funding Acquisition).

\section*{Competing interests}
The authors declare no competing interests relevant to this research project.


\section*{Ethics statement}
The use of human cadaveric specimens in this study was approved and registered with the Human Tissue Authority (HTA) Licensing Number $12778$, granted under Section $16$ ($2$) (d) of the Human Tissue Act $2004$. All specimens were obtained through a licensed body donation programme with informed consent for research use.

\section*{Open Access} 
Open Access This article is licensed under a Creative Commons
Attribution 4.0 International License, which permits use, sharing, adaptation, distribution and reproduction in any medium or format, as long as you give appropriate credit to the original author(s) and the source, provide a link to the Creative Commons licence, and indicate if changes were made. The images or other third-party material in this article are included in the article’s Creative Commons licence, unless indicated otherwise in a credit line to the material. If material is not included in the article’s Creative Commons licence and your intended use is not permitted by statutory regulation or exceeds the permitted use, you will need to obtain permission directly from the copyright holder. To view a copy of this licence, visit \url{http://creativecommons.org/licenses/by/4.0/}

\end{document}